\pdfoutput=1
\documentclass[11pt]{article}

\usepackage[final]{acl}

\usepackage{times}
\usepackage{latexsym}

\usepackage[T1]{fontenc}
\usepackage[utf8]{inputenc}

\usepackage{microtype}

\usepackage{inconsolata}

\usepackage{graphicx}

\usepackage{microtype}
\usepackage{booktabs} % for professional tables

\usepackage{hyperref}

\usepackage{amsmath}
\usepackage{amssymb}
\usepackage{mathtools}
\usepackage{amsthm}

\usepackage{colortbl}

\usepackage{multirow,array,tabularx, float}
\usepackage{booktabs} % For \toprule, \midrule, and \bottomrule
\usepackage{afterpage}
\usepackage{subcaption}
\usepackage{enumitem} % Add this package at the beginning

\definecolor{Gray}{gray}{0.95}

\usepackage[capitalize,noabbrev]{cleveref}

\theoremstyle{plain}

\theoremstyle{definition}

\usepackage{mathtools}
\usepackage{multirow}
\usepackage{wrapfig}
\usepackage{url}
\usepackage{xcolor}
\usepackage{enumitem}
\usepackage{capt-of}
\usepackage{booktabs}
\usepackage{subcaption}
\usepackage{calrsfs}
\newcommand{\llama}{\textsc{\mbox{LLaMa}}\xspace}
\newcommand{\opt}{\textsc{OPT}\xspace}

\usepackage{xspace}
\usepackage{soul}
\usepackage[noabbrev]{cleveref}

\newcommand{\eg}{{\it e.g.}, }
\newcommand{\ie}{{\it i.e.}, }

\usepackage{graphicx}
\graphicspath{ {./images/} }

\usepackage[textsize=tiny]{todonotes}

\title{Decoding Speculative Decoding}

\author{Minghao Yan\thanks{Correspondence: Minghao Yan <\href{mailto:myan@cs.wisc.edu}{myan@cs.wisc.edu}>}, \quad Saurabh Agarwal, \quad  Shivaram Venkataraman \\
        Department of Computer Sciences \\
        University of Wisconsin-Madison}
\begin{document}
\maketitle
% \vspace{-2cm}
\begin{abstract}
Speculative Decoding is a widely used technique to speed up inference for Large Language Models (LLMs) without sacrificing quality. When performing inference, speculative decoding uses a smaller \emph{draft} model to generate speculative tokens and then uses the \emph{target} LLM to verify those draft tokens. The speedup provided by speculative decoding heavily depends on the choice of the draft model.
In this work, we perform a detailed study comprising over 350 experiments with \llama-65B and \opt-66B using speculative decoding and delineate the factors that affect the performance gain provided by speculative decoding. 
Our experiments indicate that the performance of speculative decoding depends heavily on the latency of the draft model, and the draft model's capability in language modeling does not correlate strongly with its performance in speculative decoding. 
Based on these insights we explore a new design space for draft models and design hardware-efficient draft models for speculative decoding. 
Our newly designed draft model can provide 111\% higher throughput than existing draft models and our approach generalizes further to all \llama models (1/2/3.1) and supervised fine-tuned models.
\end{abstract}

% \vspace{-8pt}
\begin{figure*}[t]
    \centering
        \centering
        \includegraphics[width=\linewidth]{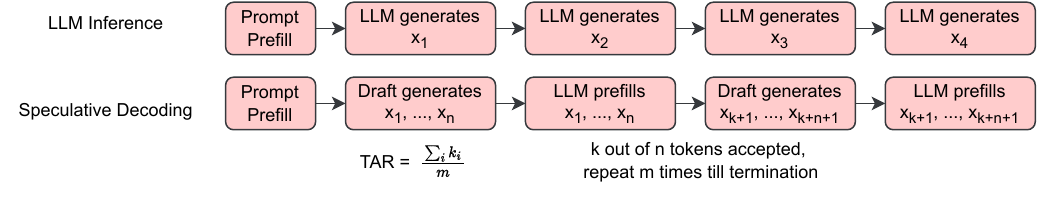}
        % \vspace{-15pt}
        \caption{This figure shows the speculative decoding process. In vanilla LLM inference, after the prompt is processed into KV caches (Prefill), LLM generates the output token by token (Autoregressive generation). In speculative decoding, a draft model is first used to generate $n$ candidate tokens at each step (Draft token generation). The LLM verifies the candidate tokens and accepts $k$ ($k \leq n$) tokens (LLM verification).} 
        % Since LLM knows all $n$ candidate tokens in advance, this step is identical to a prefill step of length $n$. In both cases, this process is repeated until either an end-of-sequence (EOS) token is generated or the maximum generation limit has been reached.}
        % \vspace{-10pt}
        \label{fig:spec_decode}
\end{figure*}
\section{Introduction}
In recent years, Large Language Models (LLMs) have emerged as a cornerstone of modern computational linguistics, offering unprecedented capabilities in generating and interpreting human language. As the demand for faster and more efficient language processing grows, understanding and optimizing the inference throughput of these models becomes increasingly crucial. Decoder-only LLMs \cite{gpt3, llama, llama2} use autoregressive decoding to perform inference. Autoregressive decoding is known to be hardware inefficient~\cite{specinfer,onlinespeculative}, leading to poor resource utilization and low throughput during inference.

Several methods~\cite{yu2022orca, wang2020lightseq, kwon2023efficient, flashdecode, hong2023flashdecoding++} have been studied to optimize the serving of LLMs. One promising approach to improve the throughput for serving LLMs without accuracy loss is speculative decoding~\cite{stern2018blockwise,xia2023speculative,leviathan2023fast}.  
% Speculative Decoding was introduced to harness the power of speculative execution to achieve a lossless speedup of translation. 
When using speculative decoding to serve an LLM (usually 10s to 100s of billion parameters), a draft model (a significantly smaller LLM) is used to generate speculative tokens.
The target LLM model then verifies the output of the draft model and only outputs tokens that match its output. In the case of speculative decoding, the target LLM for inference acts as a 
% \emph{target} 
\emph{verifier} for the draft model. 
By leveraging faster inference of smaller draft models, speculative decoding turns autoregressive decoding on the target LLM into a more hardware-friendly batched operation (similar to ``prefill''), thereby increasing throughput while preserving accuracy.

Given the promised benefits of speculative decoding, this paper first focuses on understanding the key factors that dictate the throughput improvements that can be obtained. 
%Given To improve the performance of speculative decoding,  
We perform a comprehensive benchmarking study and profile speculative decoding to characterize bottlenecks. We perform over 350 experiments, using LLMs like \llama-65B, \opt-66B, and fine-tuned chat models such as Vicuna-33B~\cite{vicuna2023} as target models and \llama and \opt families as draft models, ranging from $\approx 5\times$ to $528\times$ fewer parameters than the target models. 
Our findings show that the key bottleneck in speculative decoding is the \emph{draft model's latency} (time to generate candidate tokens from the draft model), highlighting the need to study the design of draft models for speculative decoding.

% \minghao{
While studying the design of draft models, we observed two interesting phenomena: First we observe that the draft model latency is bottlenecked by model depth, and higher model depth leads to increased latency (Section~\ref{sec:draft_latency}) even with the same number of parameters. Secondly,  we observe that draft model accuracy on language modeling tasks does not correlate strongly with its performance in speculative decoding (Section~\ref{sec:acc_vs_TAR}), \ie a draft model with higher accuracy on language modeling task can have similar TAR to a model with lower accuracy.
% Next, we find 
These two phenomena show that existing draft models used for speculative decoding were primarily designed only to achieve maximum accuracy for a given parameter budget and are sub-optimal for maximizing the throughput with speculative decoding. 

Based on these two insights, we propose designing new draft models that trade increased depth for width (thus retaining the same parameter count) and show that our new draft models can boost inference throughput using speculative decoding by over 60\%. 
Finally, we show how pruning methods like Sheared-\llama~\cite{sheared} can be used to generate smaller draft models with favorable configurations on three different families of models \opt, \llama, and \llama-3.1. 
% \todo{check}.
% \minghao{and the obtained models can be generalized to fine-tuned, newer pre-trained models in the family, and different decoding strategies.}

% We have open-sourced our code \footnote{\url{https://github.com/uw-mad-dash/decoding-speculative-decoding}} and distilled models\footnote{\url{https://huggingface.co/minghaoyan/}} on Github and HuggingFace.

\noindent \textbf{Our Contributions:}
% \vspace{-5pt}
\begin{itemize}
\item To the best of our knowledge, we are the first work to conduct comprehensive experiments on serving the open source \llama-65B and \opt-66B models utilizing speculative decoding, conducting more than 352 experiments to elucidate the factors one needs to consider while selecting and designing a draft model. 
% \vspace{-5pt}
\item We show a systematic redesign of draft models used for speculative decoding is needed to maximize the efficiency of speculative decoding. We demonstrate that using accuracy on language modeling tasks to choose the draft model for speculative decoding leads to suboptimal choices. By redesigning draft models, we improved speculative decoding throughput by up to 111\%. 
% in sampling-based decoding and 60\% in greedy decoding. 
% Based on these insights, our pruned \llama-796M provides up to $111\%$ higher throughput than Sheared-\llama-1.3B while using only $0.8\%$ of tokens (0.42B vs 50.42B) used to train Sheared-\llama-1.3B.
% We also show that \llama-796M works well for other LLMs, such as \llama-2 and 3 models, and supervised fine-tuned models (Vicuna-33B).
% \vspace{-5pt}
\item We show that our design leads to a $37\%$ reduction in KV-Caches, enabling larger batch sizes, and also outperforms other popular methods like self-speculative decoding~\cite{zhang2023draft} in several different setups.
%minghao{}
% \item Finally, we also study how improvements in models and hardware can further impact draft model design for future generations of LLMs. (Section~\ref{sec:choosing_draft_models}).
\end{itemize}
% \vspace{-10pt}

\begin{figure*}[t]
    \centering
    \begin{subfigure}[b]{0.48\linewidth}
        \includegraphics[width=\linewidth]{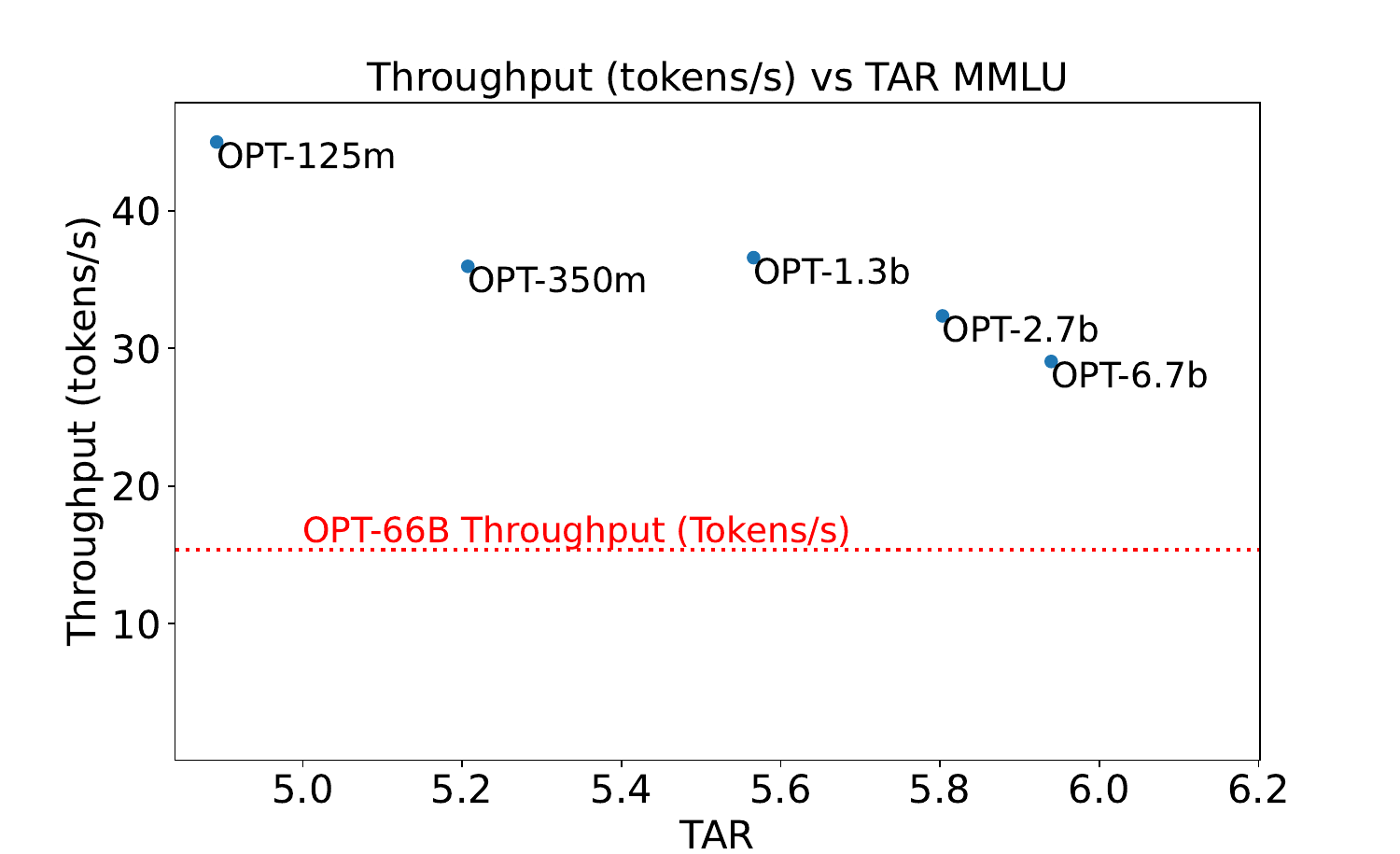}
        \caption{OPT Series on MMLU}
        \label{fig:opt_series_mmlu}
    \end{subfigure}
    % \hfill % Add some horizontal spacing
    % \begin{subfigure}[b]{0.45\linewidth}
    %     \includegraphics[width=\linewidth]{figures/tput_vs_TAR_Llama_65b_scatter_mmlu.pdf}
    %     \caption{\llama Series}
    %     \label{fig:llama_series_mmlu}
    % \end{subfigure}
    \begin{subfigure}[b]{0.48\linewidth}
        \includegraphics[width=\linewidth]{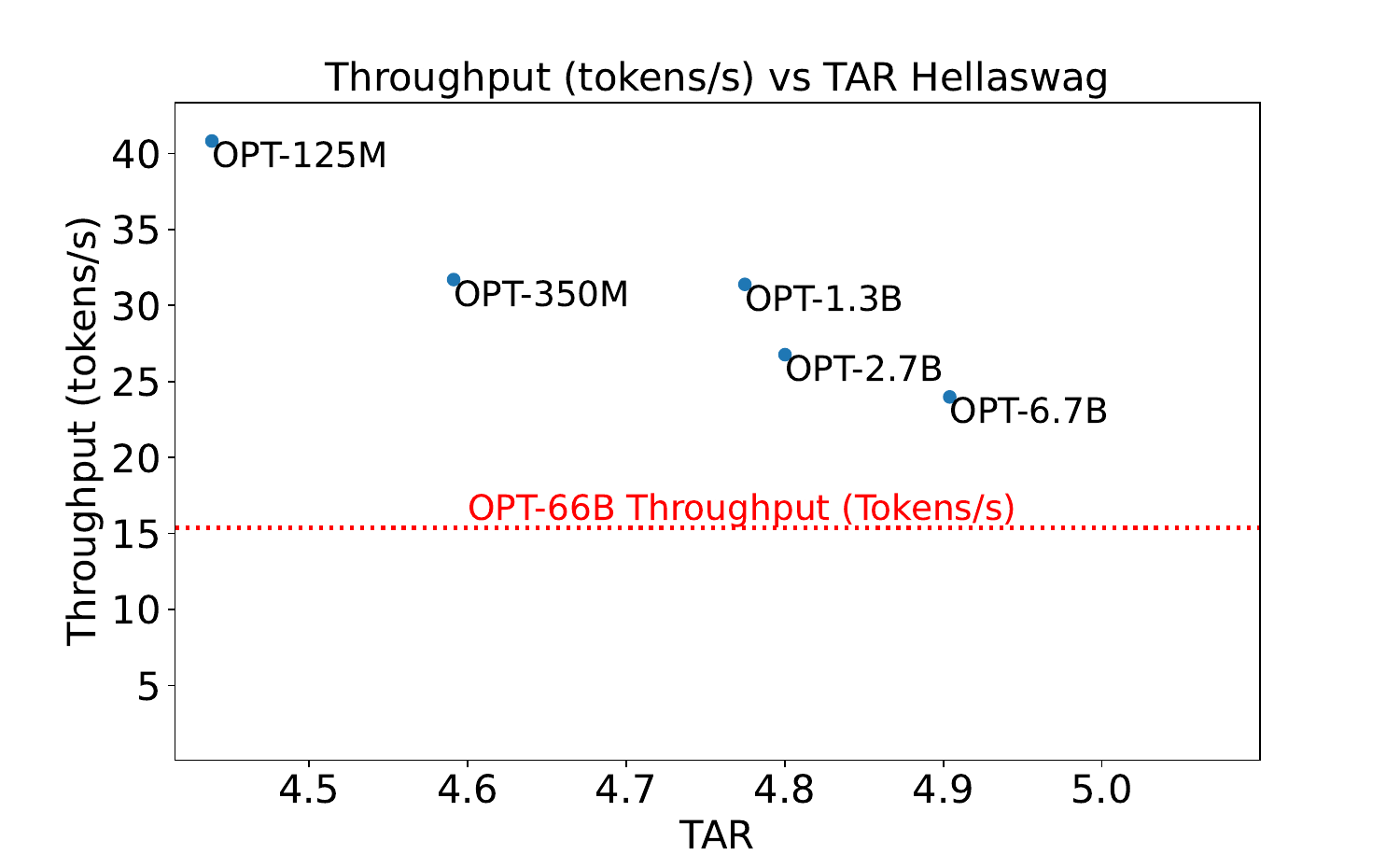}
        \caption{OPT Series on Hellaswag}
        \label{fig:opt_series_hellaswag}
    \end{subfigure}
    % \hfill % Add some horizontal spacing
    % \begin{subfigure}[b]{0.45\linewidth}
    %     \includegraphics[width=\linewidth]{figures/tput_vs_TAR_Llama_65b_scatter_hellaswag.pdf}
    %     \caption{\llama Series}
    %     \label{fig:llama_series_hellaswag}
    % \end{subfigure}
    % \vspace{-10pt}
    \caption{This figure shows the throughput of different draft models from the OPT series. As model size increases, throughput decreases due to higher inference latency despite consistent increases in TAR.}
    \label{fig:tput_vs_TAR_opt}
    % \vspace{-15pt}
\end{figure*}

% \vspace{-10pt}
\section{Background and Related Work}
First, we provide a high-level overview of LLM inference and the use of speculative decoding. 
% \vspace{-5pt}
\subsection{Background}
% \vspace{-5pt}
A decoder-only LLM performs inference in two phases: prefill and autoregressive decoding. In the prefill phase, the LLM is initialized with a context or prompt, formulated as $C = \{c_1, c_2, ..., c_n\}$, where $C$ represents the input context and $n$ the length of the prefill. In the prefill phase, the model processes the whole input context in parallel and performs next-word prediction.
During the autoregressive decoding, the model generates new text sequentially, a token at a time, building upon the context provided in the prefill phase. Due to its sequential nature, the autoregressive decoding phase is widely known to be memory bandwidth bound on modern GPUs~\cite{leviathan2023fast}. 

To improve hardware utilization and throughput, ~\cite{leviathan2023fast} and~\cite{chen2023accelerating} proposed \emph{speculative decoding}, where a smaller \emph{draft} model generates multiple tokens, and the \emph{target} LLM verifies the generated tokens in parallel. The verification is akin to \emph{prefill} stage in LLM inference. As long as more than one token is accepted on average, speculative decoding can potentially provide speedups. Figure~\ref{fig:spec_decode} shows how inference using speculative decoding differs from autoregressive decoding.
It is widely reported~\cite{specinfer, onlinespeculative} that the number of tokens accepted by the target model affects the speedup provided by speculative decoding. 
% The closer the draft model's predictions align with the target model, the faster the overall process.

In this work, we conduct a comprehensive empirical study to identify the performance bottleneck of speculative decoding and identify strategies to design the best draft model for a given LLM.

% \vspace{-5pt}
\subsection{Related Work}
\noindent \textbf{LLM Inference:} There has been significant amount of work on improving LLM serving including work in Orca~\cite{yu2022orca}, LightSeq~\cite{wang2020lightseq}, DeepSpeed Inference~\cite{aminabadi2022deepspeed}, PagedAttention~\cite{kwon2023efficient}, FlashDecoding~\cite{flashdecode} and FlashDecoding++~\cite{hong2023flashdecoding++}. These works seek to improve LLM inference by better utilization of hardware. Other lines of work have looked at pruning LLMs based on input context to speed up inference~\cite{liu2023deja} or using shallower and wider neural networks for machine translation~\cite{kasai2020deep}. However, in this work, we focus on speculative decoding~\cite{leviathan2023fast, chen2023accelerating, santilli2023accelerating}, which has been inspired by speculative execution in hardware~\cite{hennessy2011computer}.

\textbf{Speculative Decoding:} Several prior works have studied ways to improve speculative decoding. ~\citet{onlinespeculative} seeks to continuously train the draft model on the output of the target model to improve the token acceptance rate. However, training on the same hardware during inference can be challenging. Predictive Pipeline Decoding (PPD)~\cite{yang2023predictive} introduced the use of early exit~\cite{schuster2022confident} from the target model to obtain draft tokens.
Similar to PPD,  Draft\&Verify~\cite{zhang2023draft} seeks to combine the use of early exit with speculative decoding, where the early exit~\cite{schuster2022confident, bae2023fast} from the target model acts as a draft token. 
A drawback of these methods is that the maximum benefit in latency is capped. For example, in speculative decoding, we can use draft models that are orders of magnitude (e.g., $\approx$100x-1000x) smaller than the target model, while early exit methods usually exit after performing inference over at least a fourth of the model~\cite{schuster2022confident}, thus, limiting the gain in throughput. Other lines of work, such as Medusa~\cite{cai2024medusa}, propose fine-tuning multiple generation heads within the LLM that do not match the LLM output distribution exactly but maintain the generation quality. In addition, other works have looked at improving speculative decoding via learning an encoder-decoder-based draft model~\cite{xia2023speculative}, generating multiple draft tokens~\cite{sun2024spectrfast, yang2024multi}, draft model distillation~\cite{zhou2024distillspec}, or focusing on long-context scenarios~\cite{sun2024triforce, chen2024magicdec}.
% \minghao{}
% \vspace{-1pt}

% In this work, we aim to understand the performance bottleneck of 
In this work, we aim to understand how the choice of draft model affects the throughput provided by speculative decoding. We use insights from benchmarking to design draft models that maximize speculative decoding throughput.

\begin{figure}[!h]
    \centering
    % First Subfigure
    % \vspace{-5pt}
    \begin{subfigure}{\linewidth}
        \includegraphics[width=\linewidth]{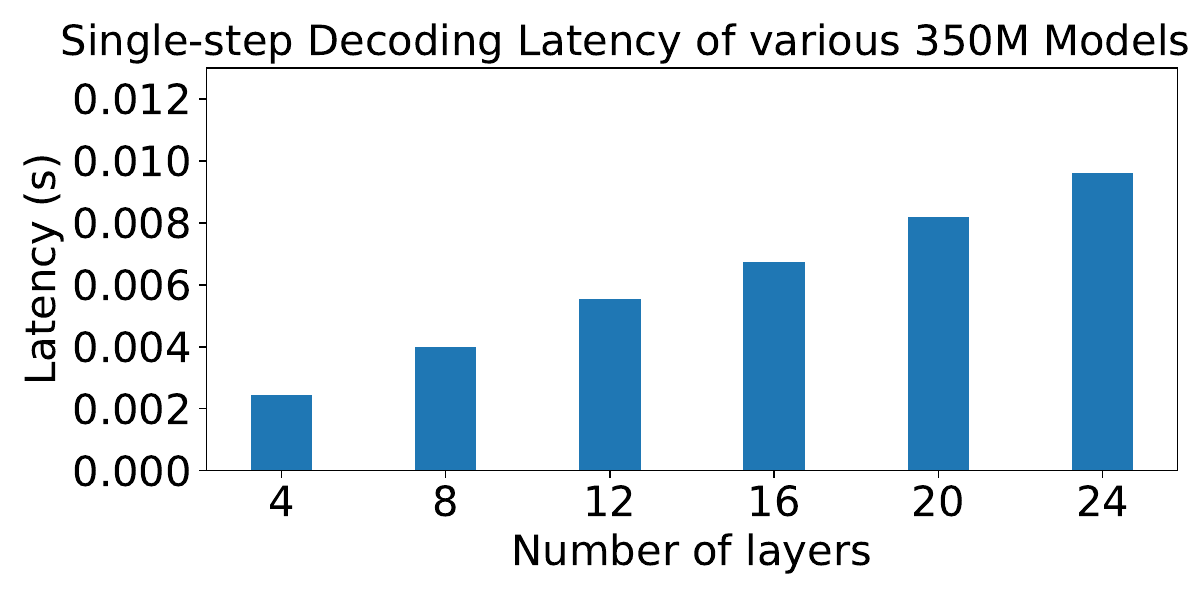}
        \caption{In this figure, we fix model parameters to 350M and vary the number of layers and attention heads. As the number of layers decreases from 24 to 4, the number of attention heads increases from 16 to 56 (Table~\ref{tab:350m_configs} in the Appendix).}
        \label{fig:latency_budget}
    \end{subfigure}

    % Second Subfigure
    \begin{subfigure}{\linewidth}
        \includegraphics[width=\linewidth]{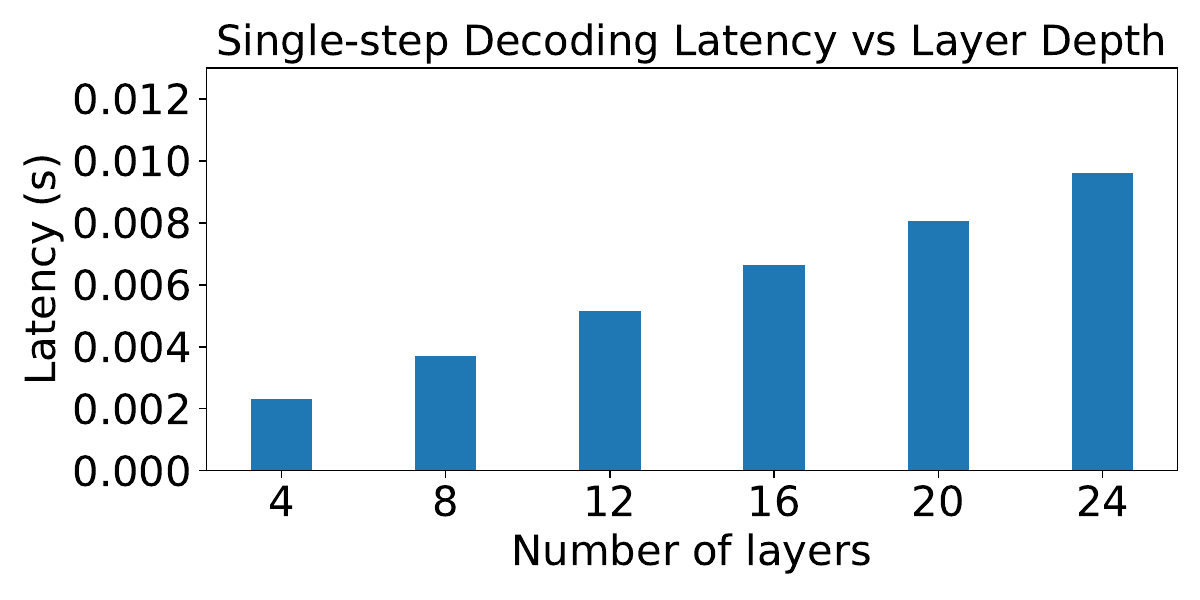}
        \caption{In this figure, we fix layer width and increase the number of layers. The number of parameters in the model increases from 79M to 350M.}
        \label{fig:latency_depth}
    \end{subfigure}

    % Third Subfigure
    \begin{subfigure}{\linewidth}
        \includegraphics[width=\linewidth]{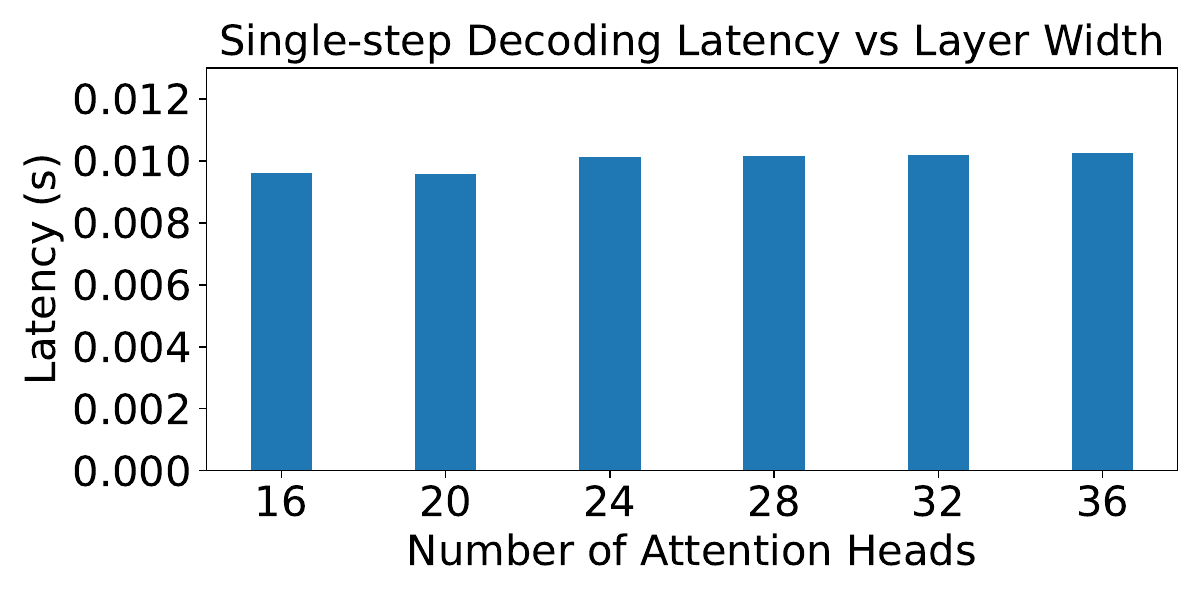}
        \caption{In this figure, we fix model depth and increase the number of attention heads in each layer. The number of parameters in the model increases from 350M to 1B.}
        \label{fig:latency_width}
    \end{subfigure}

    \caption{This figure shows microbenchmarks on how model depth and width affect decoding latency.}
    \label{fig:latency_scaling_350m}
    % \vspace{-15pt}
\end{figure}

% \vspace{-8pt}
\section{Understanding Speculative Decoding} 
\label{sec:decoding_speculative_decoding}
% \vspace{-2pt}
To study the effects of the choice of the draft model, we first perform a detailed study on serving \opt-65B and \llama-65B using speculative decoding.

\textbf{Setup:} We implement speculative decoding in the Microsoft Deepspeed library~\cite{deepspeed}. We use the same setup as SpecInfer~\cite{specinfer}, first using the draft model to generate draft tokens and then using the target model to verify the output of the draft model. We set the batch size to 1 and use greedy decoding. 
For all our experiments, we use 4 Nvidia 80GB A100 GPUs. We perform our experiment on the \opt and \llama base models~\cite{opt, llama} on MMLU~\cite{mmlu}, Hellaswag~\cite{hellaswag}, and Chatbot Arena datasets~\cite{lmsys}. For MMLU, we use the standard 5-shot setup. 
The remaining datasets were evaluated in a zero-shot setting. Note that since our goal is to test our draft model's ability to study the target model's behavior, we do not instruct the model to emit a single-letter answer to MMLU and Hellaswag questions, but instead opt for an open-ended generation approach where the model can provide as much as explanation as it sees fit, which aligns better with a real-world chatbot setting. For MMLU, we feed the model with the question and choices; for Hellaswag, we feed the model with incomplete sentences without the choices. We use \opt-66B and \llama-65B as the target LLM for OPT and \llama series and use \opt-125M, 350M, 1.3B, 2.7B, and 6.7B variants as draft models for OPT series, and \llama-7B and 13B as draft models for \llama series.

\textbf{Metrics:} To quantify the performance of different draft models when performing inference on a target model, we measure throughput (tokens generated per second) and TAR (Figure~\ref{fig:tput_vs_TAR_opt}). We note that the primary goal of speculative decoding is to improve throughput. 
% \vspace{-5pt}

\begin{figure}[!h]
    \centering
        \centering
        % \vspace{-10pt}
        \includegraphics[width=\linewidth]{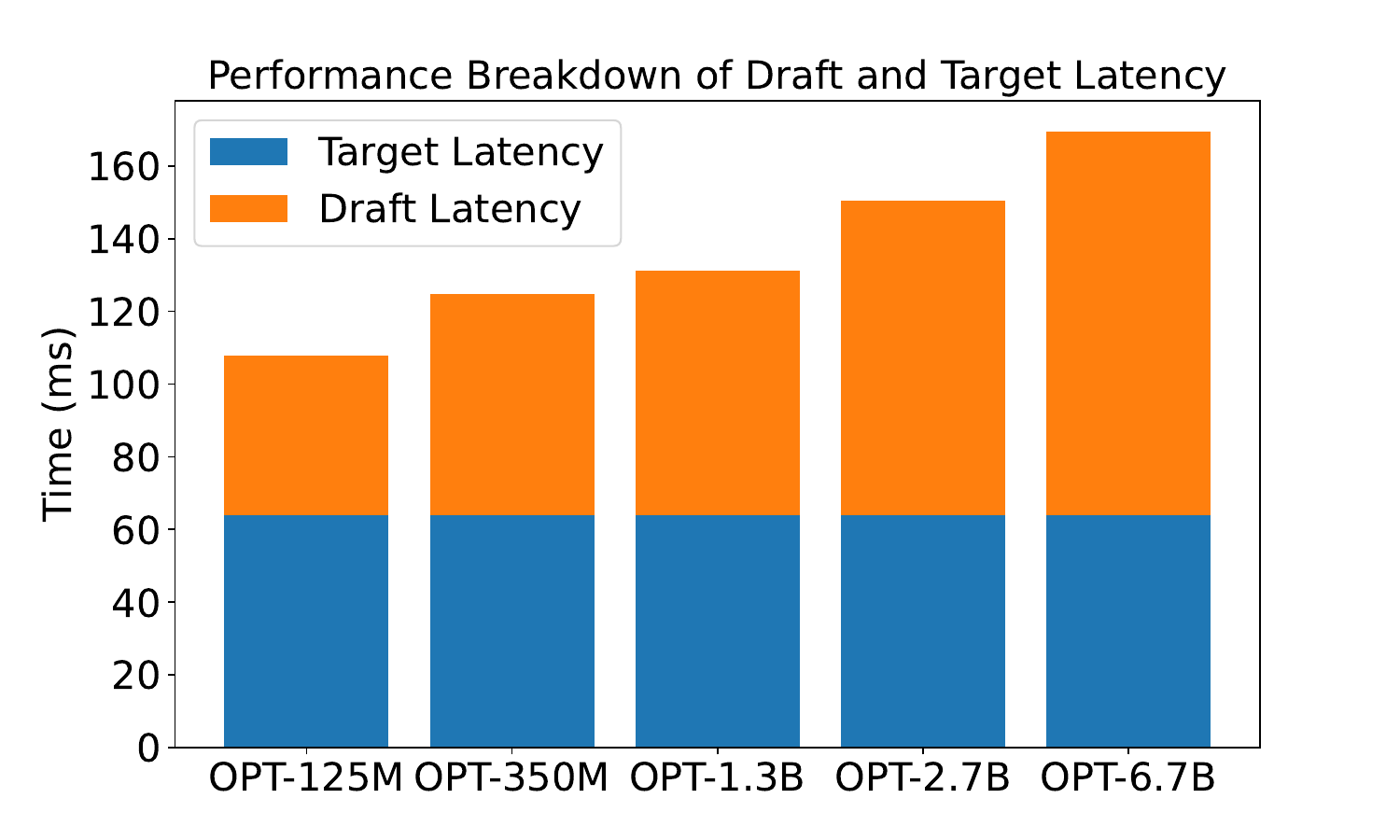}\
        % \vspace{-8pt}
        \caption{This figure shows the performance breakdown of speculative decoding on \opt models, look ahead length is set to be optimal for each draft model found empirically.}

        \label{fig:latency_breakdown}
        % \vspace{-8pt}
\end{figure}

\begin{figure*}[htp]
    \centering
    % First Row
    % \begin{minipage}{0.48\textwidth}
    %     \centering
    \begin{subfigure}[b]{0.48\linewidth}
    \includegraphics[width=\textwidth]{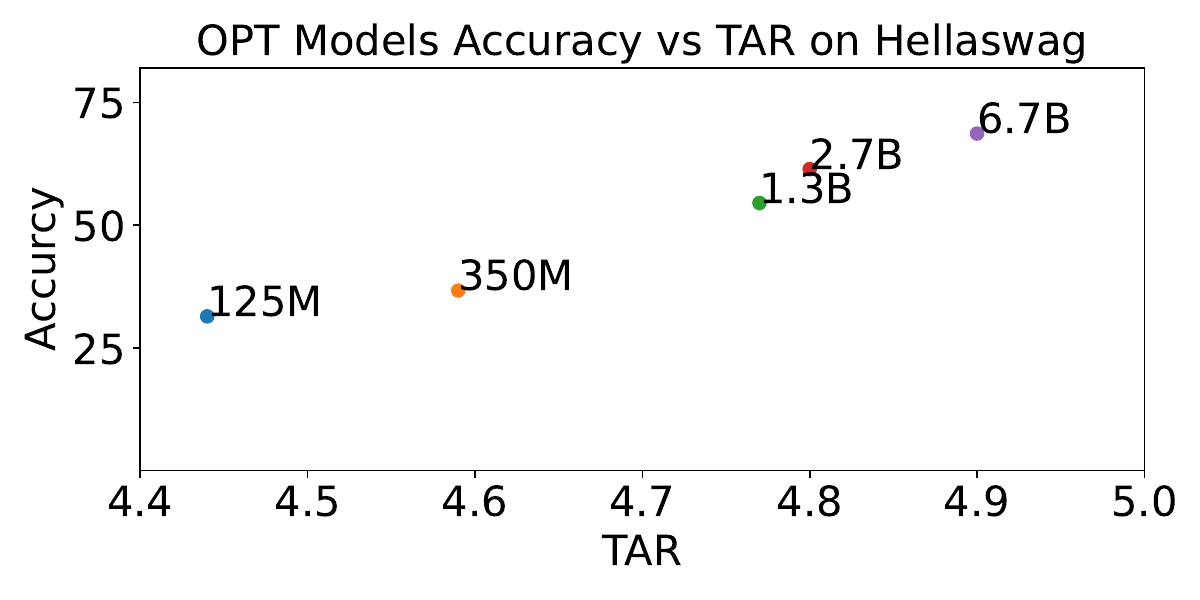}
     % \vspace{-20pt}
    \caption{Model accuracy vs TAR for OPT models}
     % \vspace{-10pt}
    \label{fig:acc_vs_tar_opt}
    \end{subfigure}
    \begin{subfigure}[b]{0.48\linewidth}
    \includegraphics[width=\textwidth]{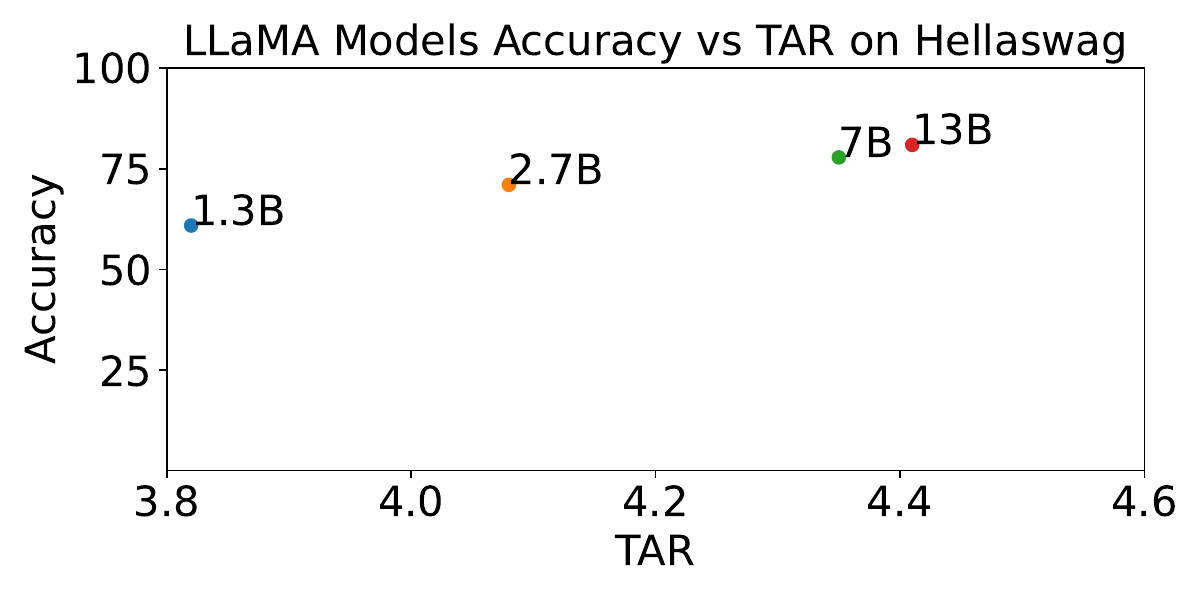}
     % \vspace{-20pt}
    \caption{Model accuracy vs TAR for \llama models}
     % \vspace{-10pt}
    \label{fig:acc_vs_tar_llama}
    \end{subfigure}
    % \end{minipage}
    \hfill
    \caption{This figure shows the task accuracy versus TAR for OPT and \llama models on Hellaswag. The accuracy numbers are obtained from OpenLLM Leaderboard~\cite{LLMLeaderboard}.}
    \label{fig:acc_vs_tar_hellaswag}
    % \vspace{-5pt}
\end{figure*}
% \vspace{-10pt}
\subsection{Bottlenecks in Speculative Decoding}\label{sec:model_depth}

% \vspace{-5pt}

To understand the throughput of LLMs, we first plot a latency breakdown of speculative decoding in Figure~\ref{fig:latency_breakdown}. We show the latency breakdown between the draft token generation phase and the target model verification phase for serving \opt-66B model when using various variants of \opt as the draft model. A similar figure for \llama models (Figure~\ref{fig:llama_perf_breakdown}) can be found in the Appendix. 
% \minghao{

In Figure~\ref{fig:latency_breakdown}, the time taken by the draft model for each token generation step increases with an increase in model sizes, going from 6.23 ms for OPT-125M to 18.56 ms for OPT-6.7B (Table~\ref{tab:draft_step_latency} in the Appendix). However, even the smallest draft model, OPT-125M, still takes significant time in a speculative decoding iteration to perform draft model autoregressive decoding. Though the target LLM has a higher latency in each decoding iteration, it only has to perform one prefill operation on the entire candidate token sequence. In contrast, the draft model has to perform multi-step autoregressive decoding sequentially, creating a bottleneck. This highlights why draft model latency is one of the key bottlenecks in speculative decoding performance. We note that while Figure~\ref{fig:latency_breakdown} uses look ahead values (the number of tokens generated by the draft model) from 6 to 8, depending on the draft model, even if we scale look ahead length to hundreds of tokens, the target model verification time stays constant. The draft model latency remains the bottleneck due to the difference in efficiency between prefill and autoregressive decoding.
Next, we investigate how to reduce draft model latency. 

% \vspace{-5pt}
\subsection{Understanding Draft Model Latency} \label{sec:draft_latency}

When studying the breakdown in latencies for speculative decoding in the previous section, we observed something intriguing in Figure~\ref{fig:latency_breakdown}. We see that OPT-350M has a similar draft-model latency as OPT-1.3B, a model almost four times its size. This indicates that OPT-350M is inefficient, and we can design better models.

We perform three microbenchmarks to validate our hypothesis and analyze decoding throughput: First, we fix the total model parameters at 350M and see how changing layer width and depth would affect decoding latency. Then, we fix either the layer width or depth to be the same as in OPT-350M and modify the other to see how latency scales with wider layers or shallower models.

Figure~\ref{fig:latency_scaling_350m} shows the results of these three benchmarks. In the first benchmark (Figure~\ref{fig:latency_budget}), we vary the number of attention heads, feed-forward dimension, and layers in a model to keep the model parameters at around 350M. The detailed configuration for each model can be found in Table~\ref{tab:350m_configs} in the Appendix. The plot shows that autoregressive decoding latency scales linearly with layer depth despite similar total model parameters.

The same is true for the second benchmark (Figure~\ref{fig:latency_depth}). The original OPT-350M model has 24 layers. As we reduce the number of layers while keeping all other configurations the same, the autoregressive decoding latency decreases linearly. On the other hand, the third benchmark (Figure~\ref{fig:latency_width}) shows that as we scale the number of attention heads up from the original OPT-350's 16 heads to 36 heads, the decoding latency stays almost constant even if layer width has doubled. 

These experiments indicate more latency-efficient model architectures with the same parameter budget exist. Changing the number of layers and attention heads not only changes the throughput but also affects the quality of predictions made by the model. We will next study how changes in model depth and width affect model accuracy and TAR and the correlation between them.

% \vspace{-8pt}
\subsection{Understanding Draft Model TAR} \label{sec:acc_vs_TAR}
% \vspace{-3pt}
In prior work~\cite{leviathan2023fast}, speculative decoding throughput is modeled by $\frac{1-\alpha^{\gamma + 1}}{(1-\alpha)(\gamma c + 1)}$, where $\frac{1-\alpha^{\gamma + 1}}{1-\alpha}$ represents the improvement factor (expected number of tokens matched in each iteration) and $\gamma c + 1$ represents the combined latency of draft and target models. Therefore, tokens accepted per iteration (also known as TAR) have a linear effect on speculative decoding throughput.

In this section, we perform experiments to understand the correlation between the accuracy of a model on popular NLP tasks and its TAR. We plot the accuracy of a model against the TAR it achieves in Figure~\ref{fig:acc_vs_tar_hellaswag}. Surprisingly, we find that TAR correlates little to the model's accuracy on a task. We believe this lack of correlation is due to the majority of tokens in a sentence not being content words \cite{chen2023accelerating}, which do not affect the model's accuracy on a specific task. Results on more datasets can be found in the Appendix (Figure~\ref{fig:acc_vs_tar_mmlu}).

% For example, if a user asks a model: \textit{What is the capital of Uruguay?} An LLM may correctly answer: \textit{The capital of Uruguay is Montevideo.} But a draft model, without retaining this much knowledge, may respond incorrectly: \textit{The capital of Uruguay is Paris.} For model accuracy evaluation, this would be a failure. However, this would be a good set of candidate tokens in speculative decoding, as the first five words are generated correctly. Therefore, as shown in Figure~\ref{fig:acc_vs_tar_hellaswag}, TAR increases sub-linearly with an increase in model size, irrespective of its accuracy on the task. 

% We discuss additional modeling choices and how to incorporate other works in Section~\ref{sec:model_discussion}.
Combining insights from these experiments, we observe that current draft models are not designed to maximize speculative decoding throughput. Next, we will show how to design new draft models that outperform existing models.

% \vspace{-8pt}
\section{Draft Model Design for Speculative Decoding}
% \vspace{-3pt}
The above results indicate that to improve the throughput of speculative decoding, it is necessary to improve the latency of draft models, 
\ie can we design a model that provides a similar TAR at a lower inference cost? In the next section, we study the possibility of such a design.
% based on the above insights. 

% \vspace{-8pt}
\begin{table}[h]
 \centering
 \caption{This table shows the model configuration of the two pruned models. Here l represents the number of layers, h is the number of attention heads, $d_{\text{inter}}$ is intermediate size, and $d_{\text{model}}$ is model dimension.}
 \small % Make the font of the entire table smaller
 \setlength{\tabcolsep}{3.2pt} % Reduce padding
\begin{tabular}{ lcccc } 

 \toprule
\rowcolor{Gray} Model & l & h & $d_{\text{inter}}$ & $d_{\text{model}}$ \\
 \midrule
 NoFT-1.3B & 24 & 16 & 5504 & 2048 \\ 
 NoFT-Wide-1.3B & 12 & 20 & 9280 & 2560 \\
 NoFT-Wide-796M & 5 & 32 & 11008 & 4096 \\
 NoFT-Wide-543M & 3 & 32 & 11008 & 4096 \\
 NoFT-Wide-290M & 1 & 32 & 11008 & 4096 \\
  \bottomrule
 \end{tabular}
 \label{tab:draft_model_config}
% \vspace{-10pt}
\end{table}

\begin{figure*}[t]
    \centering
    % \hfill % Add some horizontal spacing
    \begin{subfigure}[b]{0.48\linewidth}
        \includegraphics[width=\linewidth]{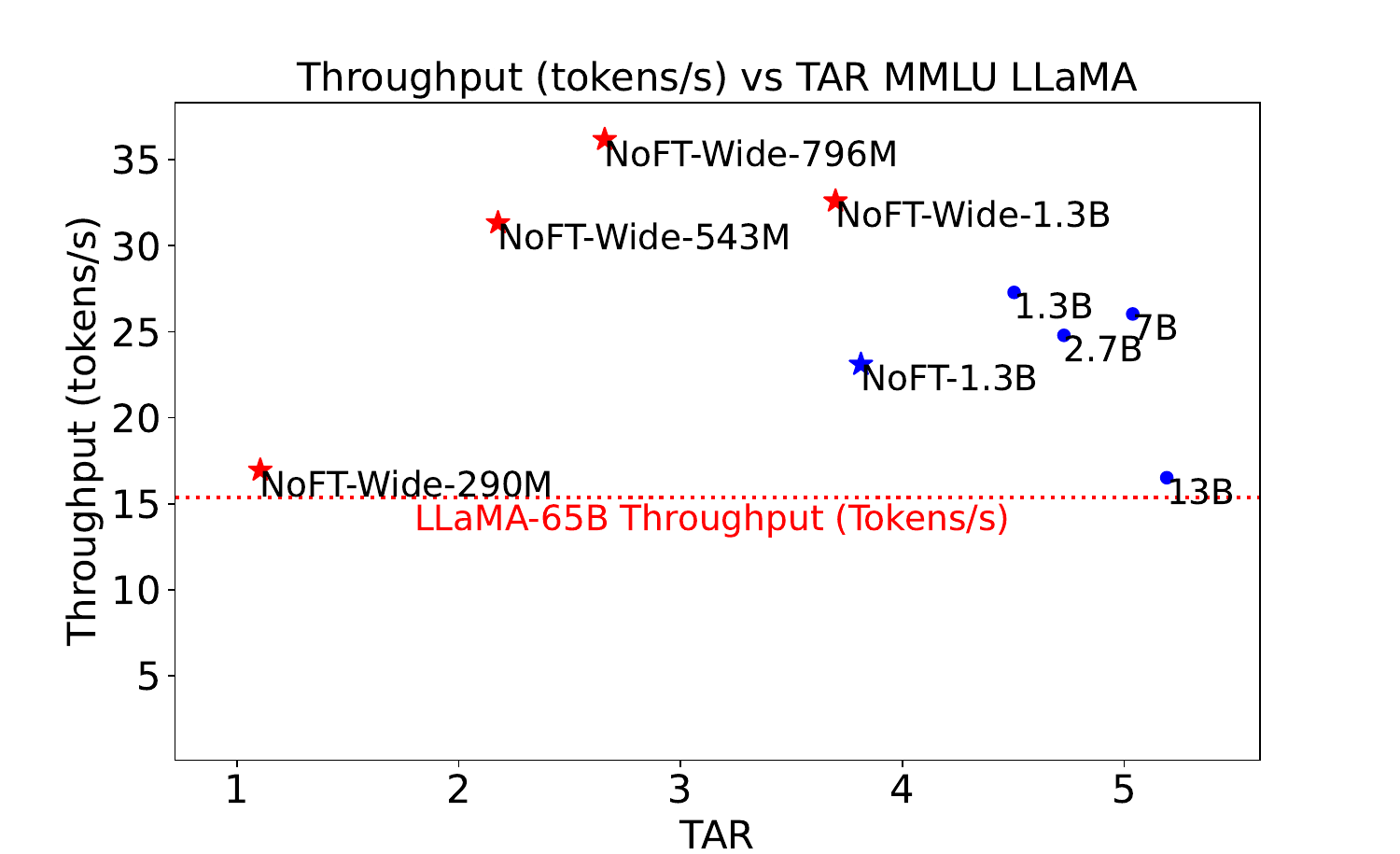}
        % \vspace{-10pt}
        \caption{\llama Series on MMLU}
        \label{fig:llama_series_mmlu}
    \end{subfigure}
    % \hfill % Add some horizontal spacing
    \begin{subfigure}[b]{0.48\linewidth}
        \includegraphics[width=\linewidth]{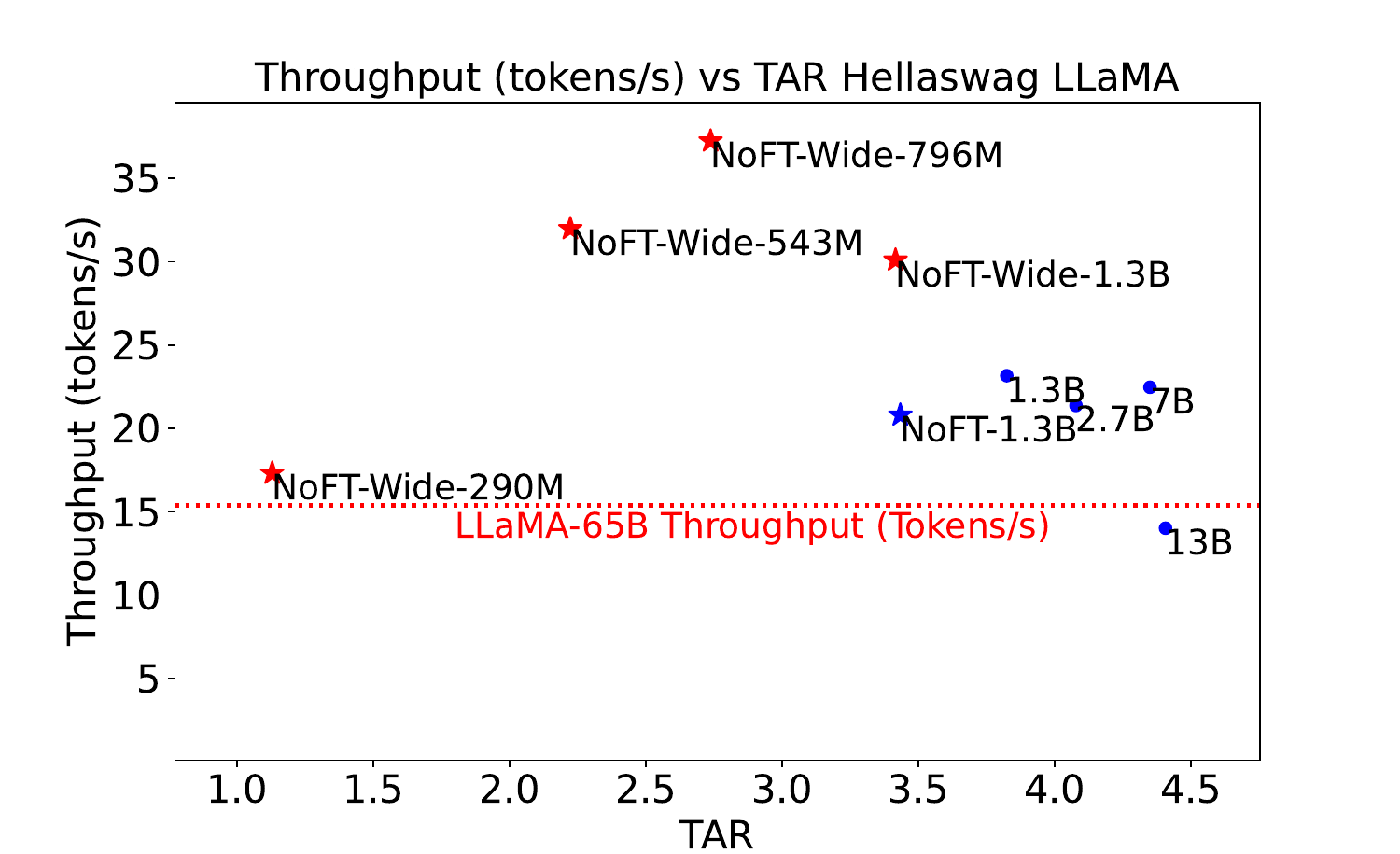}
        % \vspace{-10pt}
        \caption{\llama Series on Hellaswag}
        \label{fig:llama_series_hellaswag}
    \end{subfigure}
    % \vspace{-10pt}
    \caption{This figure shows the throughput scaling of different draft models from the \llama series on MMLU and Hellaswag. Asterisks represent models that are pruned but not fine-tuned. The red asterisks represent model configurations that we designed.}
    \label{fig:tput_vs_TAR_llama}
    % \vspace{-5pt}
\end{figure*}

\begin{table*}[t]
 \centering
 \small % Make the font of the entire table smaller
 \setlength{\tabcolsep}{3.2pt} % Reduce padding
 \caption{This table shows the throughput of speculative decoding (tokens/s) with temperature sampling on various datasets and models. Our NoFT-Wide-796M model achieved $97.7\%$ to $111.8\%$ higher throughput compared to the existing Sheared-LLaMA-1.3B model while only using $0.8\%$ of its training tokens.}
% \vspace{-5pt}
\begin{tabular}{ lcccccc } 
 \toprule
 & \multicolumn{3}{c}{Temperature = 1.0} & \multicolumn{3}{c}{Temperature = 0.5} \\
 \rowcolor{Gray} Draft Model  & MMLU & Hellaswag & Chatbot Arena & MMLU & Hellaswag & Chatbot Arena \\
 \midrule
 NoFT-Wide-543M & 20.54 & 24.08 & 23.77 & 26.08 & 25.77 & 25.31 \\ 
 NoFT-Wide-796M & \textbf{24.32} & \textbf{24.32} & \textbf{24.18} & \textbf{29.64} & \textbf{26.91} & \textbf{25.76} \\
 NoFT-Wide-1.3B & 22.99 & 24.08 & \textbf{24.22} & 28.43 & 26.39 & \textbf{25.85} \\
 Sheared-LLaMA-1.3B & 12.06 & 12.30 & 11.45 & 14.60 & 13.45 & 12.66 \\ 
 \midrule
 Speedup & 101.6\% & 97.7\% & 111.8\% & 103.0\% & 100.0\% & 103.5\% \\
 \bottomrule
 \end{tabular}
 \label{tab:vary_draft_model_tab}
% \vspace{-10pt}
\end{table*}

\subsection{Draft Model Design} \label{sec:model_design}

In section~\ref{sec:model_depth}, we show that model depth bottlenecks draft model latency, while in section~\ref{sec:acc_vs_TAR}, we show that a draft model's performance in speculative decoding is largely irrelevant to its accuracy on language modeling. These two insights prompted us to test if we can build a wider and shallower network and study how it affects latency and TAR.

\textbf{Method:} We leverage recent advances in structured LLM pruning, Sheared-\llama~\cite{sheared}, which provides a framework to prune larger models to a specified smaller configuration. Sheared-\llama~\cite{sheared} learns layers, attention heads, and neurons to mask from the large model to prune it into the specified small model. The flexibility enables us to prune \llama-7B into desirable model configurations. In our experiments, we pruned our models from \llama-7B using 0.4B tokens sampled from the RedPajama Dataset~\cite{redpajama} following \citet{sheared} but skipped the expensive fine-tuning step on 50B more tokens (and hence the name NoFT). We find that this is sufficient to achieve a significantly higher throughput. 

% \shivaram{say why we chose this 1.3B. Something like it was the best in the previous exp?}

\begin{table}[h!]
 \centering
 \caption{This table shows the TAR, per-step decoding latency, and speculative decoding throughput using the two pruned draft models.}
 \small
\begin{tabular}{ lccc } 
 \toprule
\rowcolor{Gray} Draft Model  & TAR & Latency & Throughput \\
\rowcolor{Gray} &  &  (ms) & (tokens/s) \\
 \midrule
 NoFT-1.3B  & \textbf{3.81} & 13.13 & 23.10  \\ 
 NoFT-Wide-1.3B & 3.70 & \textbf{6.69} & \textbf{32.59}   \\
  \bottomrule
 \end{tabular}
 \label{tab:draft_deep_vs_wide}
% \vspace{-10pt}
\end{table}

\textbf{Deep vs wide model comparison:} Our goal is to start with \llama-7B and produce a wider version of Sheared-\llama-1.3B while keeping the number of parameters the same as in Sheared-\llama-1.3B. We choose Sheared-\llama-1.3B since it achieves the highest throughput in our benchmark among existing models (blue dots in Figure~\ref{fig:tput_vs_TAR_llama}). We use two configurations: the first configuration was provided by the Sheared-\llama authors (NoFT-1.3B), and we designed the second configuration (NoFT-Wide-1.3B) to optimize for better speculative decoding throughput. Table~\ref{tab:draft_model_config} shows the detailed configuration of the two models. We slash the number of layers by half, from 24 to 12, and keep the total parameter count roughly the same by increasing the intermediate size from 5504 to 9280, the number of attention heads from 16 to 20, and the corresponding model dimension from 2048 to 2560. Figure~\ref{fig:llama_series_mmlu},~\ref{fig:llama_series_hellaswag}, and~\ref{fig:tput_vs_TAR_llama_chat} show that we can achieve up to $30\%$ higher speculative decoding throughput using only $0.8\%$ of tokens used to train Sheared-\llama-1.3B.

Table~\ref{tab:draft_deep_vs_wide} also shows the latency and TAR of the two sheared models on MMLU. The deep variant (NoFT-1.3B) can achieve $3\%$ higher TAR, but the wide variant (NoFT-Wide-1.3B) reduces draft latency by $49\%$, improving overall throughput by $41\%$. We found results are very similar for other datasets, such as Chatbot Arena (Figure~\ref{fig:tput_vs_TAR_llama_chat} in the Appendix) and Hellaswag (Figure~\ref{fig:llama_series_hellaswag}).
This experiment shows a need to rethink the model design space for speculative decoding, where we should specifically design models for higher throughput.

% Our model also reduces intermediate memory usage by $37\%$ compared to the original model. The memory saving comes from wider MLP layers, which do not require KV-caches.

\textbf{Draft model scaling:} To understand the limitation of draft model depth-width tradeoff in speculative decoding, we created three configurations, NoFT-Wide-796M, 543M, and 290M, that use the same number of attention heads, intermediate size, and model dimension as \llama-7B, but reduce the number of layers to 5, 3, and 1, respectively. This is the widest configuration possible using the Sheared-\llama pruning scheme. 

%\textbf{Results:}  

Figure~\ref{fig:tput_vs_TAR_llama} shows that the NoFT-Wide-796M model provides another $20\%$ improvement in throughput over NoFT-Wide-$1.3$B and up to 60$\%$ throughput improvement over the existing Sheared-\llama-$1.3$B model. Though the smaller NoFT-Wide-543M provides up to $40\%$ throughput improvements over Sheared-\llama-$1.3$B, it has a lower throughput than NoFT-Wide-796M. 

Results in Figure~\ref{fig:tput_vs_TAR_llama} show that reducing the layer count to less than 5 layers would cause the model's alignment capability to reduce dramatically. In addition, as we reduce models to 5 layers, target model latency takes more than 80$\%$ of the time in a decoding cycle. Therefore, further reducing the latency would only provide a marginal gain in overall decoding latency since the target model latency remains constant. In this case, the drop in TAR significantly outweighs the latency gain, causing decoding throughput to decrease. 

% \textbf{Temperature sampling: }
% \minghao{
We also verified our model's robustness under different temperatures. We performed experiments with temperature sampling by choosing temperatures of 0.5 and 1 following~\citet{leviathan2023fast}. In Table~\ref{tab:vary_draft_model_tab}, we show that our model achieves even higher speedup than existing models when stochasticity is introduced to the decoding process. On three datasets we achieved from $97\%$ to $111\%$ higher throughput using our NoFT-Wide-796M model compared to using an existing fine-tuned model. The speedups in different temperature settings show that our speedups generalize across different sampling methods.
% }
% the sampling-based decoding setting versus the greedy-based decoding setting shows that speedups achieved from our proposed 

\begin{table}[h]
\setlength{\tabcolsep}{2pt}
\centering
 \small
 \caption{This table shows the memory usage of our draft model \llama-Wide-1.3B versus Sheared-\llama-1.3B. Our draft model reduces KV cache by $37\%$.}
\begin{tabular}{c|cc|cc}
\toprule
 & \multicolumn{2}{c}{KV cache} & \multicolumn{2}{c}{Activations} \\

\rowcolor{Gray} Context Length & Sheared & Wide & Sheared & Wide \\
% \rowcolor{Gray} Length & \llama & Wide & \llama & Wide\\
\midrule
256  & 48MB  & \textbf{30MB}  & 1MB   & 1.25MB  \\
512  & 96MB  & \textbf{60MB}  & 2MB   & 2.5MB   \\
1024 & 192MB & \textbf{120MB} & 4MB   & 5MB     \\
\bottomrule
\end{tabular}
\label{tab:kvcache}
% \vspace{-5pt}
\end{table}

\begin{table}[h]
\setlength{\tabcolsep}{2pt}
\centering
 \small
 \caption{This table shows the prefill and per-step autoregressive decoding latency for different batch sizes.}
\begin{tabular}{cc|cc}
\toprule
Batch size & Prefill (s) & \multicolumn{2}{c}{Autoregressive decoding (s)} \\

\rowcolor{Gray} Model  & LLaMA-65B & LLaMA-NoFT & Sheared \\
\rowcolor{Gray}   & & Wide-1.3B & LLaMA-1.3B \\
\midrule
1  & 0.060  & \textbf{0.0097} & 0.018 \\
4  & 0.061  & \textbf{0.0097} & 0.019 \\
16 & 0.063  & \textbf{0.0099} & 0.019 \\
32 & 0.065  & \textbf{0.0101} & 0.019 \\
\bottomrule
\end{tabular}
\label{tab:scale_bs}
% \vspace{-10pt}
\end{table}
% \vspace{-5pt}

\textbf{Batch size and KV cache:} By designing a shallower model with wider MLP layers, we reduce the KV-cache size required by the model. This allows us to increase the batch size we can accommodate. Table~\ref{tab:kvcache} shows that for the same sized 1.3B model, our wider design reduces KV-Cache by more than $37\%$. Table~\ref{tab:scale_bs} demonstrates that increasing the batch size has minimal impact on autoregressive decoding latency, thereby increasing throughput.
% }

\textbf{\llama-3 Results} \label{app:llama3} We apply our approach to the newest \llama-3.2-1B model. All other experiment settings follow section~\ref{sec:model_design}. We evaluated on MMLU, Hellaswag, and Chatbot Arena prompts. We compare our Wide-829M model against \llama-3.2 and self-speculative decoding. Since \llama-3 is trained on more tokens than \llama-2 (15 trillion vs 2 trillion), pruning its layers had a larger impact on TAR. We follow prior work~\cite{muralidharan2024compact} to recover the TAR after pruning by distilling it over 1 million tokens. Note that this process can be finished within 10 minutes. In this experiment, we use \llama-3.1-8B as the target model. Table~\ref{tab:draft_llama3} shows that our pruned Wide-829M model achieves $42.6\%$ higher throughput compared to \llama-3.2-1B and $51.8\%$ higher throughput compared to self-speculative decoding. This experiment shows that our approach works well on SoTA models.
% }

\begin{table}[h]
 \centering
 \caption{This table shows the speculative decoding throughput and the latencies of our Wide-829M models against \llama-3.2-1B model and self-speculative decoding.}
 \small
\begin{tabular}{ lccc } 
 \toprule
 & \multicolumn{3}{c}{Throughput(tokens/s)} \\
\rowcolor{Gray} Draft Model  & Chat & MMLU & Hellaswag  \\
 \midrule
 Wide-829M  & \textbf{53.35} & \textbf{61.77} & \textbf{60.44}\\ 
 LLaMA-3.2-1B & 37.41 & 44.89 & 39.03 \\
 Self-Speculative & 35.14 & 40.41 & 40.02\\
  \bottomrule
 \end{tabular}
 \label{tab:draft_llama3}
% \vspace{-10pt}
\end{table}

% \vspace{-8pt}
\subsection{Ablation Studies}
% \vspace{-3pt}
In this section, we study if varying the decoding sampling methods or using a different or supervised fine-tuned target model would affect our draft model's performance.

% }

% \begin{table}[t]
%  \centering
%  \caption{This table shows the throughput of speculative decoding (tokens/s) with temperature sampling on various datasets and models.}
%   \small % Make the font of the entire table smaller
%  \setlength{\tabcolsep}{3.2pt} % Reduce padding
%    % \vspace{-5pt}
% \begin{tabular}{ lccc } 

%  \toprule
%  & \multicolumn{}{}{} & \multicolumn{}{}{}
% \rowcolor{Gray} Draft Model  & MMLU & Hellaswag & Chatbot Arena & MMLU & Hellaswag & Chatbot Arena \\
%  \midrule
%  NoFT-Wide-543M & 20.78 & 18.25 & 18.73 \\ 
%  NoFT-Wide-796M & \textbf{29.87} & \textbf{26.55} & \textbf{25.61} \\
%  NoFT-Wide-1.3B & \textbf{29.87} & \textbf{26.55} & \textbf{25.61} \\
%  Sheared-LLaMA-1.3B & 20.78 & 18.25 & 18.73 \\ 
%   \bottomrule
%  \end{tabular}
%  \label{tab:vary_draft_model_sft}
% % \vspace{-10pt}
% \end{table}

\textbf{Varying the target model:} Prior experiments are performed with \llama-65B as the target model. As newer generations of models roll out, we would like to see if our conclusion holds on newer generations of models. In this ablation study, we evaluate our best NoFT-Wide-796M model against the \llama-2-70B model. Table 
\ref{tab:vary_target_model} shows that though our NoFT-Wide-796M is distilled from \llama-7B, it can achieve a similar token acceptance rate when the target model is from \llama-2 family. This shows that our distilled model can be applied to various models based on similar training recipes and tokenizers. We also show more results on the newest \llama-3.1 and \llama-3.2 families in Appendix~\ref{app:llama3}.
% \vspace{-10pt}
\begin{table}[h]
 \centering
 \caption{This table shows the tokens accepted per iteration when we use different target models. The draft model we use is NoFT-Wide-796M.}
  \small % Make the font of the entire table smaller
 \setlength{\tabcolsep}{3.2pt} % Reduce padding
\begin{tabular}{ lccc } 

 \toprule
\rowcolor{Gray} Target Model  & MMLU & Hellaswag & Chatbot Arena \\
 \midrule
 \llama-65B & 2.66 & 2.74 & 2.61\\ 
 \llama-2-70B & 2.55 & 2.68 & 2.64 \\
  \bottomrule
 \end{tabular}
 \label{tab:vary_target_model}
% \vspace{-10pt}
\end{table}

\textbf{Supervised fine-tuned models:} Prior experiments are performed on base models to study the scaling of draft models. In practice, supervised fine-tuned models are adopted for their better instruction-following capabilities. In this section, we compare our best NoFT-Wide-796M model to Tiny-\llama-1.1B with Vicuna 33B as the target model. Note that our NoFT-Wide-796M is pruned from the base version of \llama-7B without fine-tuning. Table~\ref{tab:vary_draft_model_sft} shows that NoFT-Wide-796M outperforms Tiny-\llama-1.1B in all cases by up to $45\%$. While Tiny-\llama-1.1B has a TAR $35\%$ and $32\%$ higher than NoFT-Wide-796M on MMLU and Hellaswag, respectively, its latency is 4x higher due to having 22 layers in the model compared to NoFT-Wide-796M with merely 5 layers. This ablation study also demonstrates how speculative decoding is bottlenecked by draft model depth and that a draft model obtained from the non-fine-tuned base model, when appropriately designed, can outperform fine-tuned models. 
% \shivaram{add TAR for the two draft models.}
% \vspace{-5pt}
\begin{table}[h]
 \centering
 \caption{This table shows the throughput of speculative decoding (tokens/s) with Vicuna 33B as the target model.}
  \small % Make the font of the entire table smaller
 \setlength{\tabcolsep}{3.2pt} % Reduce padding
   % \vspace{-5pt}
\begin{tabular}{ lccc } 

 \toprule
\rowcolor{Gray} Draft Model  & MMLU & Hellaswag & Chatbot Arena \\
 \midrule
 Tiny-\llama-1.1B & 20.78 & 18.25 & 18.73 \\ 
 NoFT-Wide-796M & \textbf{29.87} & \textbf{26.55} & \textbf{25.61} \\
  \bottomrule
 \end{tabular}
 \label{tab:vary_draft_model_sft}
% \vspace{-10pt}
\end{table}
% \vspace{-8pt}

\textbf{Comparison against self-speculative decoding:} Self-speculative decoding~\cite{zhang2023draft} was proposed to leverage the base model to avoid the need to train a draft model. However, we show that we achieve significantly higher throughput with our proposed model architecture design. We pick the settings where self-speculative decoding achieves its highest speedup (\llama-2-70B as target model with greedy sampling on CNNDM and XSum datasets). However, our \llama-NoFT-Wide-796M  outperforms self-speculative decoding by up to $53\%$. This is because our \llama-NoFT-Wide-796M model's autoregressive decoding latency is only $7\%$ of the target 70B model's prefill latency. Self-speculative decoding suffers from a significant drop in drafting accuracy after dropping more than 42 layers out of 80 in \llama-2-13B~\cite{zhang2023draft}. Table~\ref{tab:self_spec_layer} shows that on a 70B target model, they would still need to compute 33 attention layers and 43 MLP layers to obtain a draft response, while our \llama-NoFT-Wide-796M only has 5 attention and MLP layers each. Therefore, we obtain a much higher throughput compared to self-speculative decoding. More comparisons on \llama-3 are presented in Appendix~\ref{app:llama3}.
% }
% \vspace{-15pt}
\begin{table}[h]
 \centering
 % \vspace{-5pt}
 \caption{This table shows the number of layers used in self-speculative decoding and our draft model.}
  \small % Make the font of the entire table smaller
 \setlength{\tabcolsep}{3.2pt} % Reduce padding
   % \vspace{-5pt}
\begin{tabular}{ lccc } 

 \toprule
 & & \multicolumn{2}{c}{Layers}\\
\rowcolor{Gray} \multicolumn{2}{c}{Model Setup} & Attention & MLP\\
 \midrule
Self-Speculative &  \llama-2-13B & 16 & 30 \\ 
&  \llama-2-70B & 33 & 43 \\
\midrule
Draft Model  & NoFT-Wide-796M & 5 & 5 \\
  \bottomrule
 \end{tabular}
 \label{tab:self_spec_layer}
% \vspace{-10pt}
\end{table}
% \vspace{-5pt}
\begin{table}[h]
 \centering
 \caption{This table shows the speedup achieved by our \llama-NoFT-Wide-796M and self-speculative decoding, respectively.}
  \small % Make the font of the entire table smaller
 \setlength{\tabcolsep}{3.2pt} % Reduce padding
   % \vspace{-5pt}
\begin{tabular}{ lcc } 

 \toprule
\rowcolor{Gray} Strategy & \multicolumn{2}{c}{Speedup} \\
 \midrule
 & CNNDM & XSum \\
Self-Speculative &  99\% & 60\%\\ 
NoFT-Wide-796M  & \textbf{131\%} & \textbf{145\%}\\
  \bottomrule
 \end{tabular}
 \label{tab:self_spec_layer}
% \vspace{-10pt}
\end{table}
% \vspace{-5pt}
\section{Conclusion}
% \vspace{-5pt}
In this work, we conduct a large-scale experimental study to understand how we can optimize the throughput of speculative decoding. Using our experiments, we outline the various factors that affect speculative decoding throughput. We observe that draft model accuracy on language modeling does not correlate strongly with its performance in speculative decoding. Further, we find that draft model latency is bottlenecked by model depth, and higher model depth increases latency. Based on these two insights, we propose new draft models pruned to align with the target model while trading model depth for width. Our proposed draft model can increase throughput by up to 111$\%$ over existing models. We find that the pruned models can be used for supervised fine-tuned target models without modification and our approach generalizes to state-of-the-art models.

% \newpage
\section*{Limitations}
Our work aims to improve the inference efficiency of LLMs by designing better draft models for speculative decoding. One limitation of our work is that we focus on empirically studying the performance bottleneck and improving lossless speculative decoding throughput, where lossless refers to preserving the target LLM’s output distribution. Broader studies on approximate efficient inference algorithms are left for future work.

% Since speculative decoding preserves the output from the LLM, our work will not amplify existing biases in LLMs. However, limiting and reducing such biases are out of the scope of this work. Furthermore, since we are making LLM generation more efficient, we believe our work will not have a significant negative environmental impact. 

\section*{Acknowledgement}
We gratefully acknowledge the support of the NSF Diamond project OAC-2311767 (Democratizing Large Neural Network Model Training for Science). We thank Kangwook Lee for his helpful suggestions during the project. This research used resources from the National Energy Research Scientific Computing Center (NERSC), a Department of Energy Office of Science User Facility, using the NERSC award DDR-ERCAP0029980. This research also used computational resources from
the NSF Cloudlab~\cite{cloudlab} facility.

\bibliography{ref}

\appendix

\section{More Experiment Results} \label{app:tput_vs_TAR}

\begin{table}[h]
 \centering
 \caption{This table shows the latency of each autoregressive generation step of the draft model.}
    \small % Make the font of the entire table smaller
 \setlength{\tabcolsep}{5pt} % Reduce padding
\begin{tabular}{ lc } 

 \toprule
\rowcolor{Gray} Model  & Latency (ms) \\
 \midrule
 OPT-125M & 6.23 \\ 
 OPT-350M & 11.74 \\ 
 OPT-1.3B & 12.64 \\
 OPT-2.7B & 16.35 \\  
 OPT-6.7B & 18.56 \\  
  \bottomrule
 \end{tabular}
 \label{tab:draft_step_latency}
% \vspace{-0.5cm}
\end{table}

\subsection{OPT analysis}

\begin{figure*}[h]
    \centering
    % First Row
    \begin{subfigure}[b]{0.48\textwidth}
        \centering
        \includegraphics[width=\textwidth]{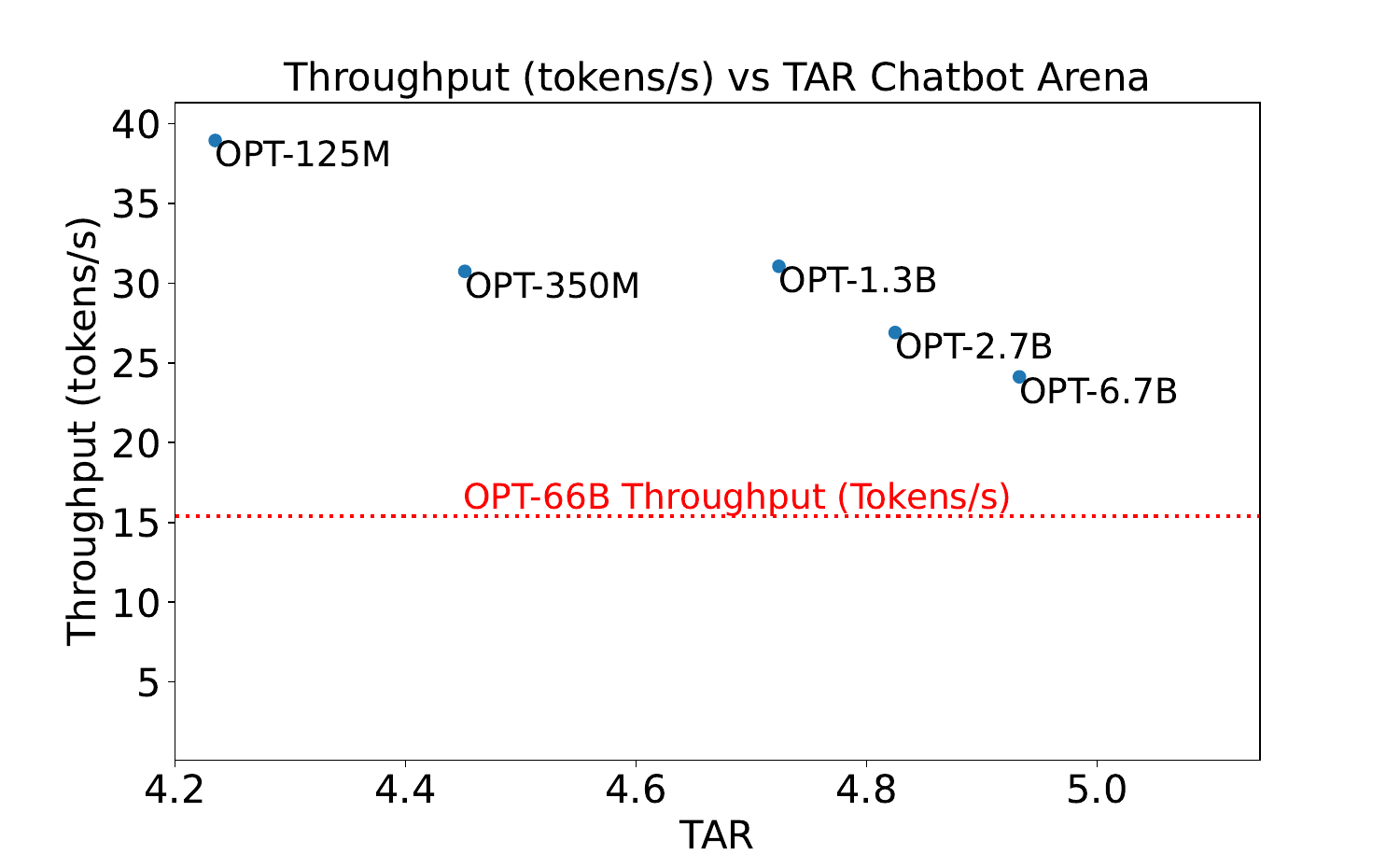}
        \caption{Throughput scaling with increasing TAR in OPT series on Chatbot Arena}
        \label{fig:tput_vs_TAR_OPT_chat}
    \end{subfigure}
    \hfill
    \begin{subfigure}[b]{0.48\textwidth}
        \centering
        \includegraphics[width=\textwidth]{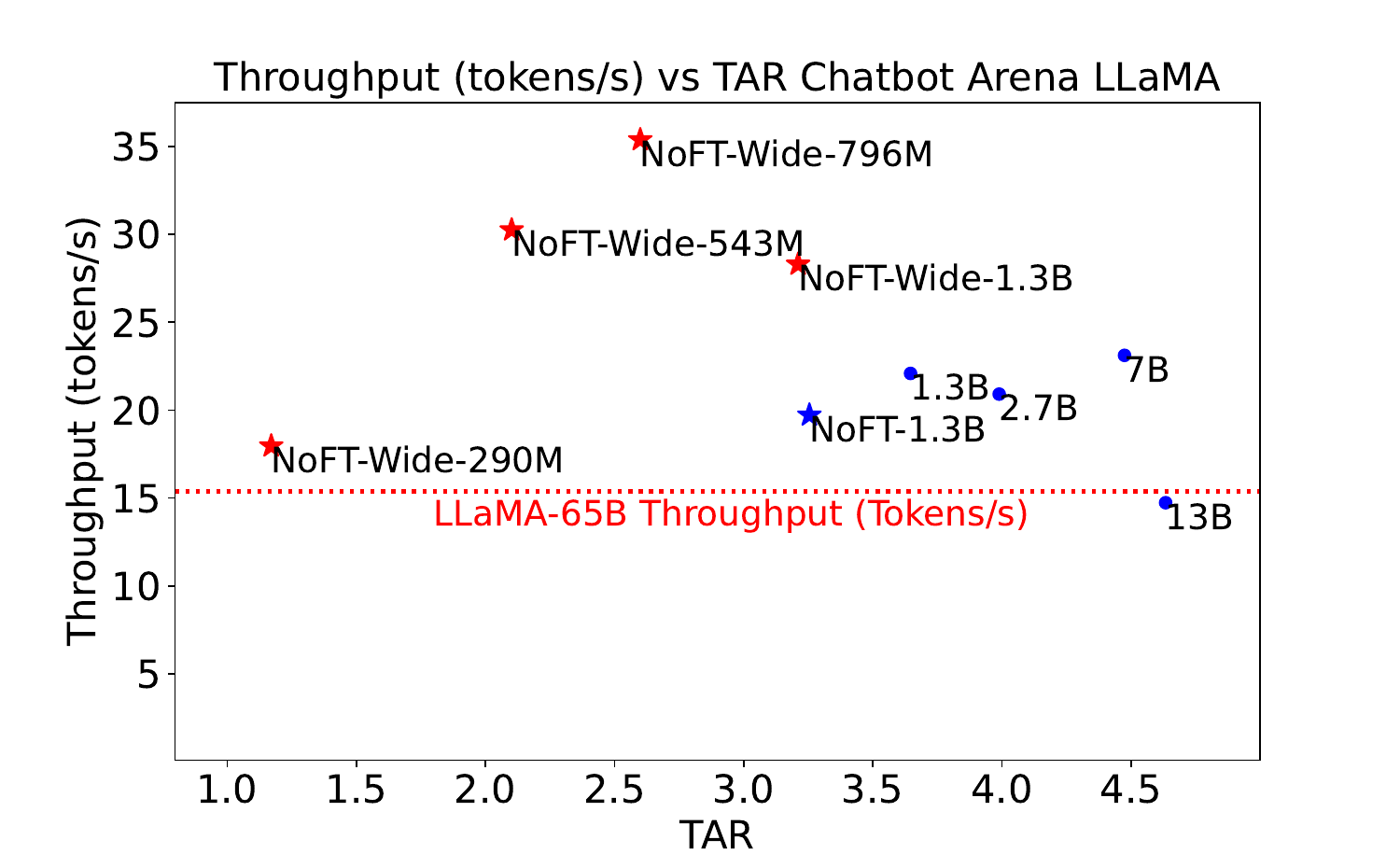}
        \caption{Throughput scaling with increasing TAR in \llama series on Chatbot Arena}
        \label{fig:tput_vs_TAR_llama_chat}
    \end{subfigure}
    \caption{This figure shows the throughput scaling against TAR for Chatbot Arena.}
    \label{fig:tput_vs_TAR_chat}
\end{figure*}

In this section, we show more experimental analysis of speculative decoding. In Figure~\ref{fig:tput_vs_TAR_chat}, we plot the throughput of \opt and \llama models against its TAR on Chatbot Arena. This figure shows that as model size increases, throughput generally decreases due to significantly higher inference latency despite consistent increases in TAR.

\begin{figure*}[h]
    \centering
    % First Row
    % \begin{minipage}{0.48\textwidth}
    %     \centering
    \begin{subfigure}[b]{0.45\linewidth}
    \includegraphics[width=\textwidth]{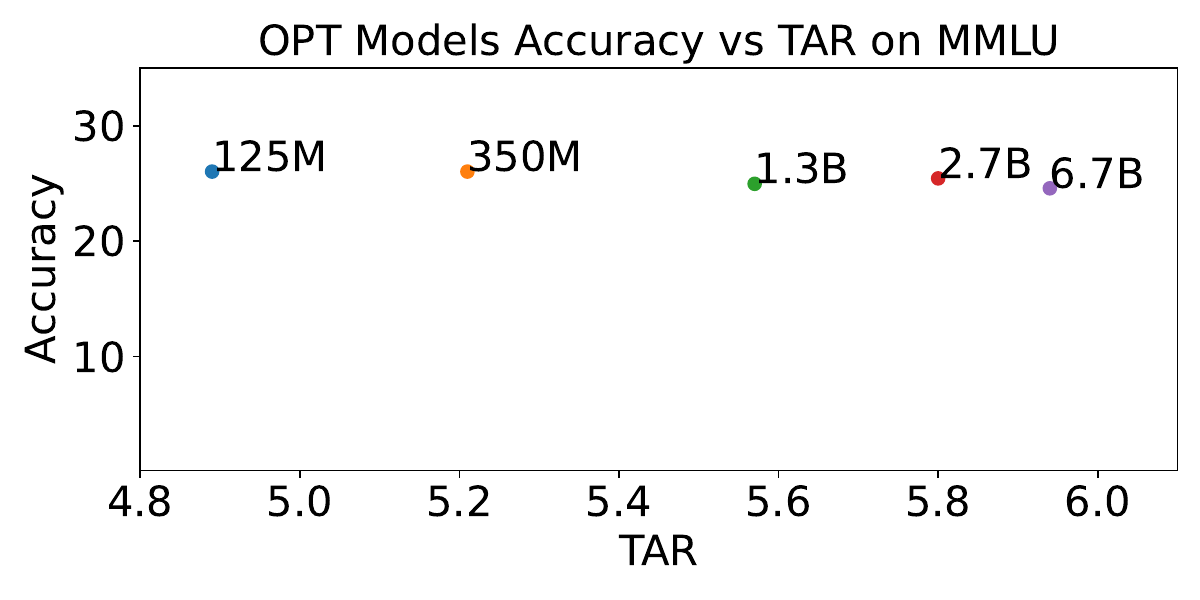}
    \caption{Model accuracy vs TAR for OPT models}
    \label{fig:acc_vs_tar_opt_mmlu}
    \end{subfigure}
    \begin{subfigure}[b]{0.45\linewidth}
    \includegraphics[width=\textwidth]{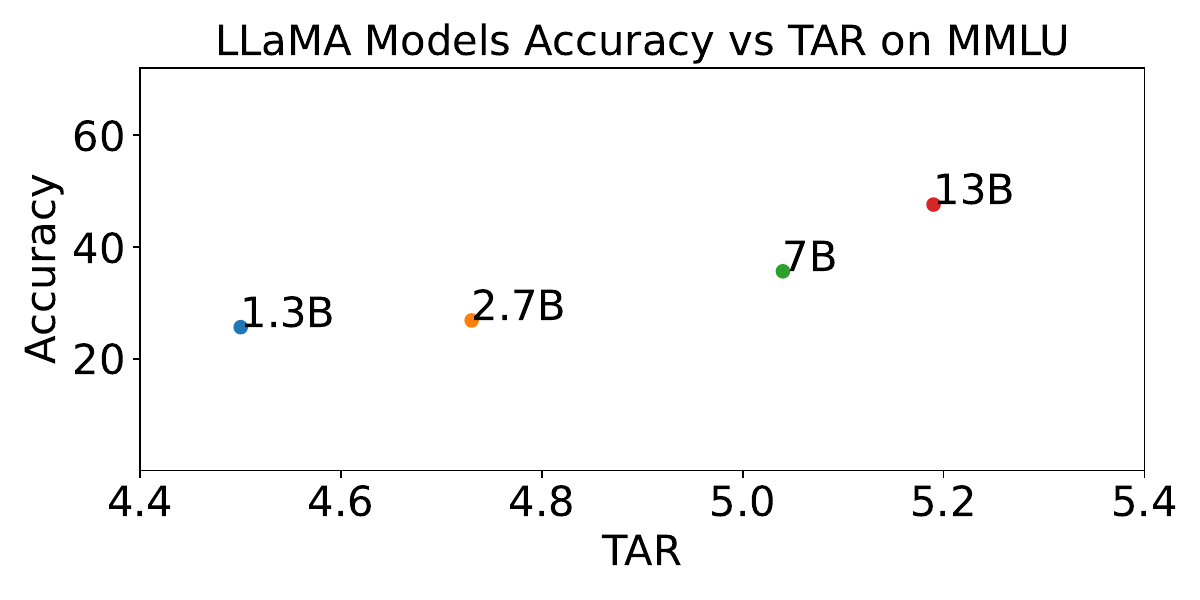}
    \caption{Model accuracy vs TAR for \llama models}
    \label{fig:acc_vs_tar_llama_mmlu}
    \end{subfigure}
    % \end{minipage}
    \hfill
    \caption{This figure shows the task accuracy versus TAR for OPT and \llama models on MMLU. The accuracy numbers are obtained from OpenLLM Leaderboard~\cite{LLMLeaderboard}.}
    \label{fig:acc_vs_tar_mmlu}
\end{figure*}

\begin{figure}[h]
    \centering
    \begin{minipage}{0.48\textwidth}
        \centering
        \includegraphics[width=\textwidth]{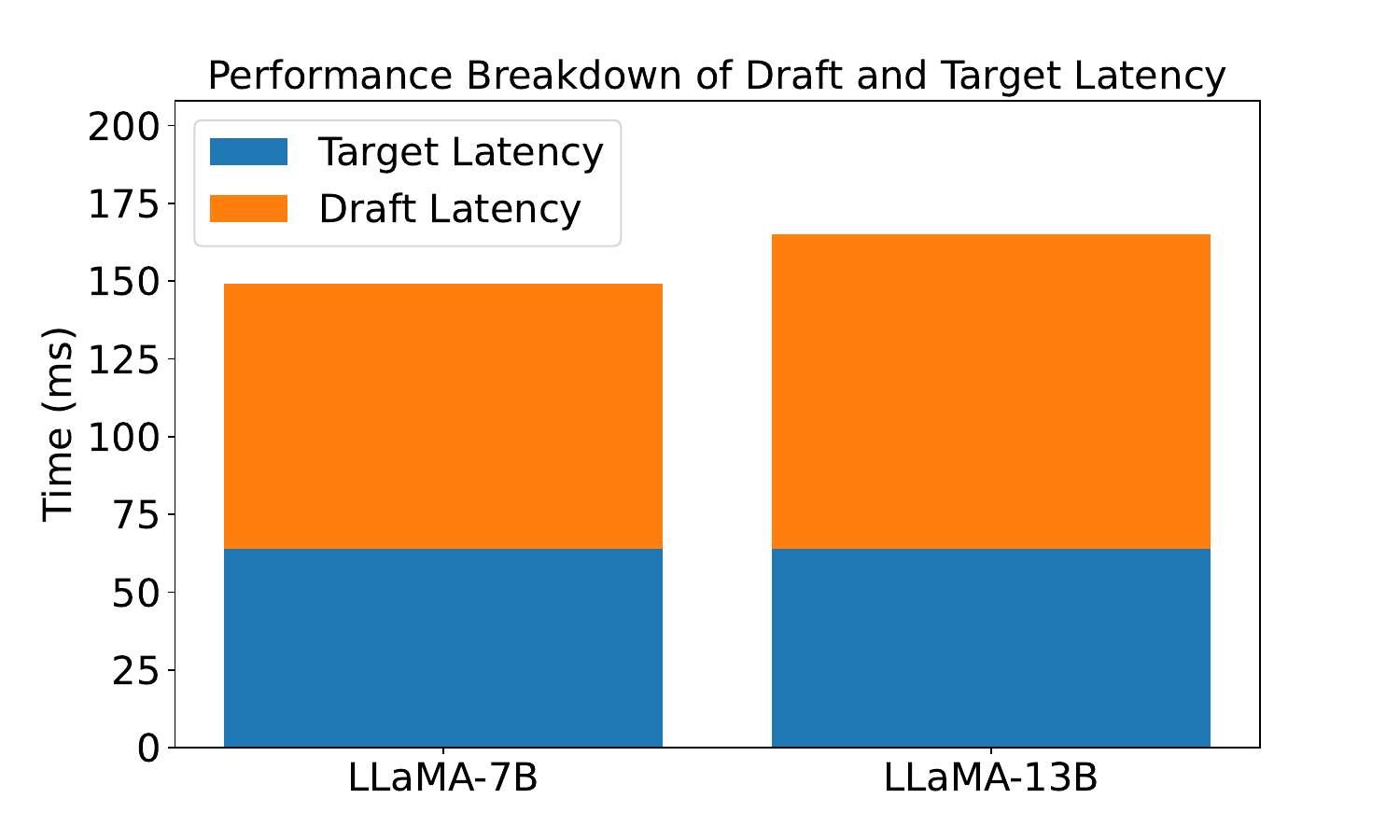}
        \caption{This figure shows the performance breakdown of \llama speculative decoding, look ahead length is set to be optimal look ahead length found empirically.}
        \label{fig:llama_perf_breakdown}
    \end{minipage}
\end{figure}

In Figure~\ref{fig:llama_perf_breakdown}, we plot the throughput of \opt and \llama models against its TAR on Chatbot Arena. This figure shows that draft latency occupies a large chunk of time in a speculative decoding iteration, opening up new avenues for designing draft models optimal for speculative decoding.
% \newpage

In Figure~\ref{fig:acc_vs_tar_mmlu}, we plot the task accuracy versus TAR for \opt and \llama models on MMLU. The accuracy numbers are obtained from OpenLLM Leaderboard~\cite{LLMLeaderboard}. This figure shows that task accuracy is irrelevant to TAR.

% \vspace{-5pt}
\section{Discussion}
% \vspace{-5pt}
In this section, we discuss how our insights can change if the models or the underlying hardware change.
% \vspace{-5pt}
\subsection{Future Draft Model Design} \label{sec:choosing_draft_models} %todo name this better. 
To study how compute and performance changes can lead to different choices of draft models, we use a performance model. The original speculative decoding~\cite{leviathan2023fast} model $\frac{1-\alpha^{\gamma + 1}}{(1-\alpha)(\gamma c + 1)}$ can be simplified to the following to remove the unnecessary assumption of mutual independence between generated tokens in a sequence:
% \vspace{-8pt}
\[
% \normalsize
\text{Throughput} =
\begin{cases}
\displaystyle \frac{TAR}{(t^{d}_{target} + t^{d}_{draft})} & \text{if } TAR > 1,\\
\displaystyle \frac{1}{(t^{d}_{target} + t^{d}_{draft})} & \text{if } TAR \leq 1.
\end{cases}
\vspace{-3pt}
\]
% \vspace{-3pt}
We show that this simplified formula almost perfectly captures the real speculative decoding throughput. Here, $t^d$ represents the latency to generate $d$ tokens autoregressively. In this section, with the aid of the performance model, we provide quantitative answers to several questions: First, we study the improvement in TAR a larger draft model needs to be provided to compensate for the additional inference cost. 
Next, we study how much improvement in latency is required to change the choice of the draft model.

\begin{table}[h]
 \centering
 \caption{This table shows the latency reduction needed for larger draft models to achieve parity throughput with OPT-125M on MMLU. }
   \small % Make the font of the entire table smaller
 \setlength{\tabcolsep}{5pt} % Reduce padding
  \vspace{-5pt}
\begin{tabular}{ lccc } 

 \toprule
\rowcolor{Gray} Model  & Latency (ms) & Parity Latency & Reduction ($\%$) \\
 \midrule
 125M & 43.7 & 43.7 & 0 \\ 
 350M & 79.8 & 50.6 & 36.6 \\ 
 1.3B & 87.1 & 58.7 & 32.6 \\
 2.7B & 114.3 & 49.8 & 56.4\\  
 6.7B & 139.5 & 68.2& 51.1\\  
  \bottomrule
 \end{tabular}
 \label{tab:parity}
% \vspace{-10pt}
\end{table}
% \vspace{-8pt}

\begin{figure}[h]
    \centering
    % First Row
        \centering
        \includegraphics[width=0.95\linewidth]{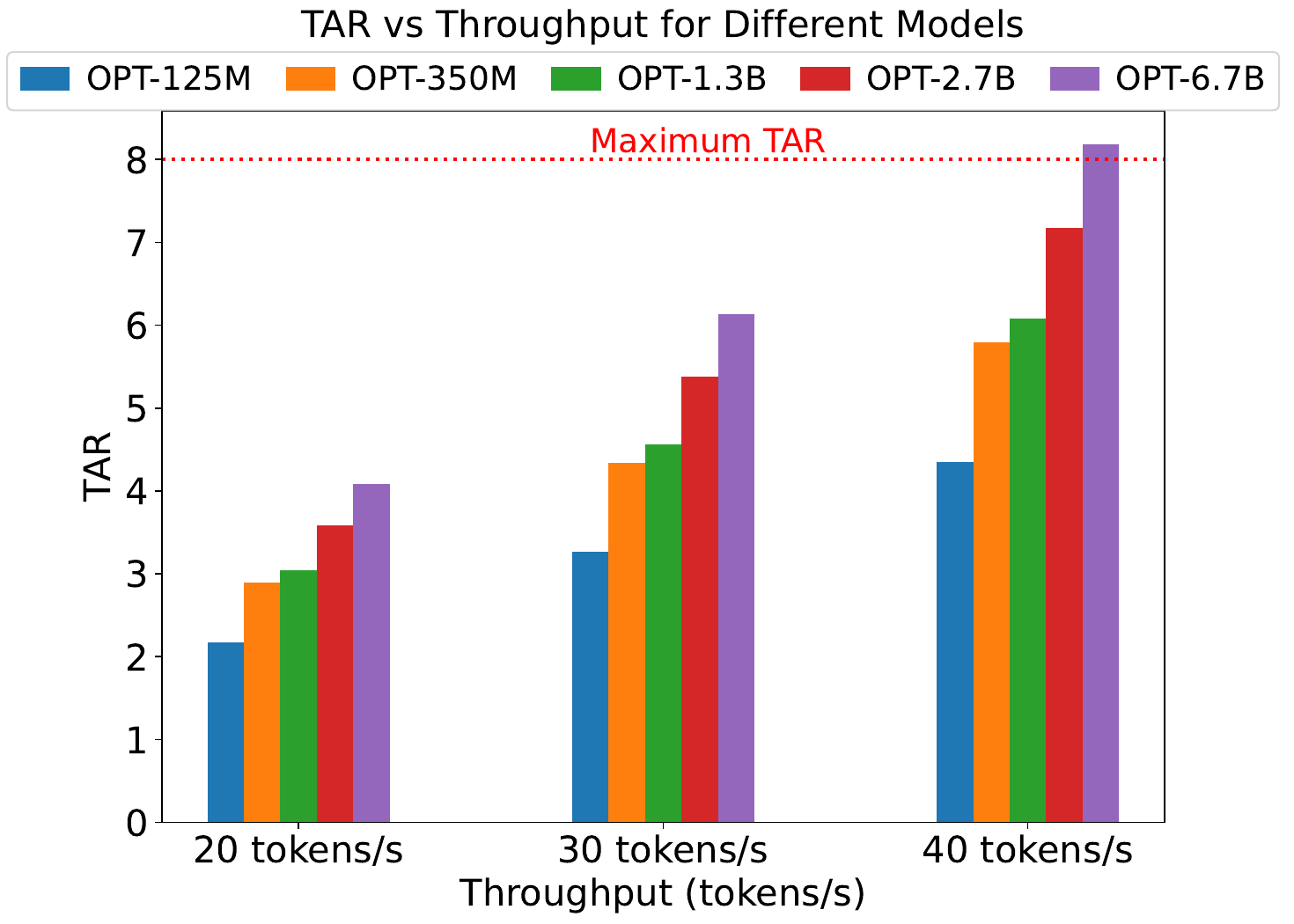}
        \caption{This figure shows the required TAR to achieve a given throughput.}
        \label{fig:opt_TAR_tput_scaling}
        \vspace{-10pt}
\end{figure}

\textbf{Required TAR to match throughput: } We use our analytical model to predict the TAR necessary for different models to achieve a target throughput. This can be useful in scenarios where developers deploy speculative decoding-based LLMs and must meet a throughput goal.
In Figure~\ref{fig:opt_TAR_tput_scaling}, we plot the TAR needed by existing models to achieve a specific throughput.

The figure shows that the TAR gap between draft models at each given throughput is much larger than we observed in Figure~\ref{fig:tput_vs_TAR_opt}. When the throughput requirement is high, a large draft model, such as \opt-6.7B, can't achieve the desired throughput. This will allow model designers to quickly judge which draft and target model pair allows them to meet throughput requirements. 

\begin{figure}[h]
    \centering
    \includegraphics[width=0.9\linewidth]{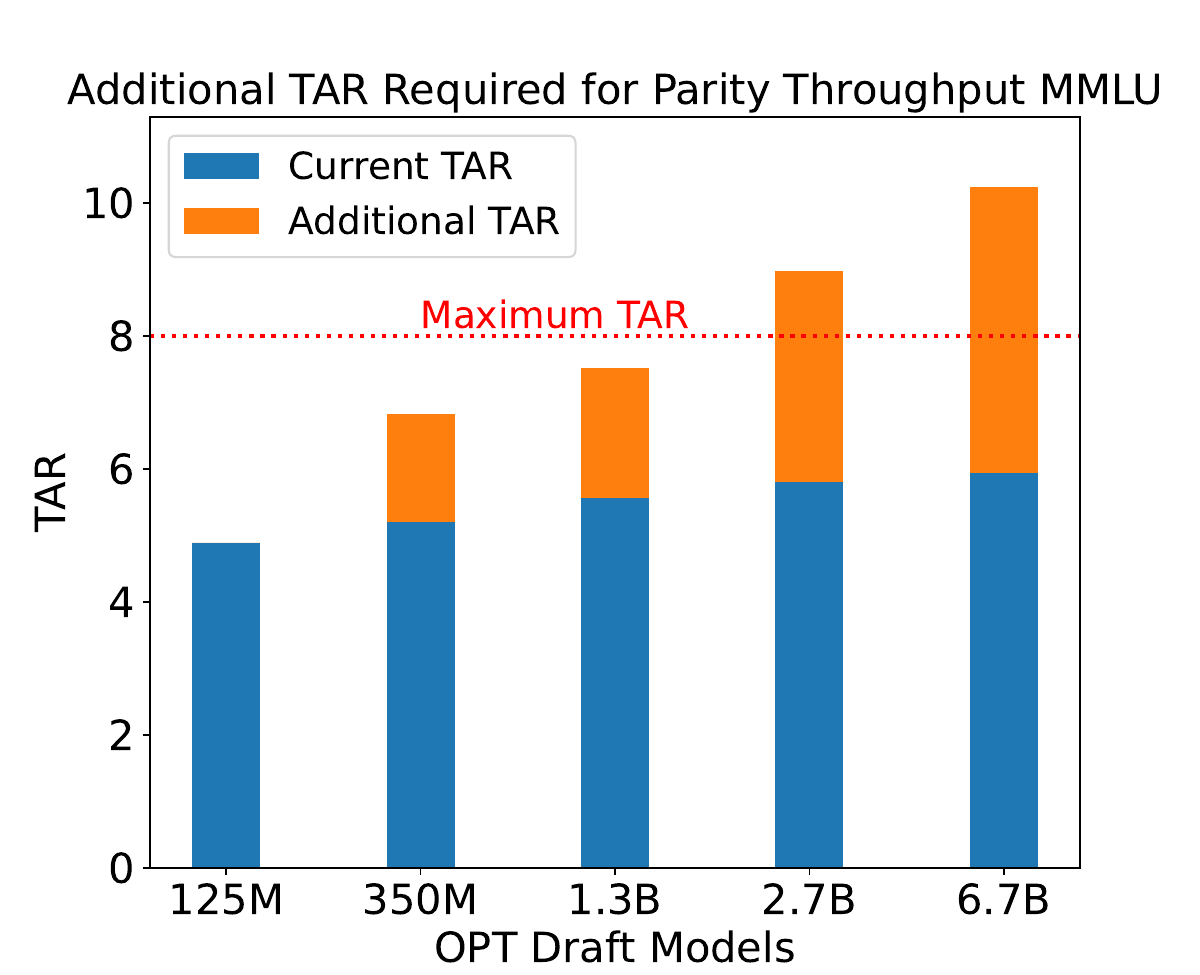}
    \caption{This figure shows the extra TAR needed for each model to achieve parity throughout with \opt-125M on MMLU.}
    \label{fig:delta_tar}
    % \vspace{-15pt}
\end{figure}

\textbf{Improvement in TAR needed to switch to a larger draft model:} In  Figure~\ref{fig:tput_vs_TAR_opt}, we observed that with existing datasets and models, we are better off with the smallest model as the draft model, \eg \opt-125M, than choosing a larger model. 
However, there is a possibility that the TAR difference will become greater for new datasets. In Figure~\ref{fig:delta_tar}, we plot the improvement in TAR (extra TAR), which larger models in the \opt model family should provide to match the throughput of the smallest model (\opt-125M) for MMLU. 
We find that if a 1.3B model can achieve a TAR advantage greater than 2 over \opt-125M for a new workload, we would choose the 1.3B model instead. Furthermore, given that the maximum TAR is capped at 8 in our scenario due to the length of draft token generation, it becomes unfeasible for \opt-2.7B and \opt-6.7B to surpass \opt-125M in performance. This is because the improvement needed in TAR for \opt-6.7B to match the throughput of \opt-125M would exceed this maximum limit.

% \vspace{-8pt}

\textbf{Improvement in latency for switching to higher TAR model:} As hardware evolves, latency scaling patterns may change with more computing power and memory bandwidth. Therefore, conclusions drawn on specific hardware (\eg A100) may not hold for newer or older hardware (\eg H100 or V100). To account for changing hardware, we study how much draft model latency improvement is needed to achieve throughput parity. 
To demonstrate this, we first compute the latency reduction needed for different members in \opt family to reach the same throughput as the smallest draft model in Table~\ref{tab:parity}. We find that up to $56\%$ of latency reduction is needed to achieve the same throughput. For instance, for \opt-1.3B to achieve parity throughput with \opt-125M, its latency needs to be reduced by 32.9\%. This reinforces our finding that latency reduction provided by the smaller models has significantly more benefit than the extra TAR provided by a larger draft model.

\section{OPT-350M Configurations}
Table~\ref{tab:350m_configs} shows the detailed model configurations of the OPT-350M variants we created. The goal is to keep the total parameter count close to that of OPT-350M while adjusting model width and depth.

\begin{table}[h]
 \centering
 \caption{This table shows the model configuration of various OPT-350M models we created. The goal is to explore the tradeoff between model depth and width while keeping the total parameter count constant.}
  \small % Make the font of the entire table smaller
 \setlength{\tabcolsep}{3.2pt} % Reduce padding
\begin{tabular}{ cccc } 

 \toprule
\rowcolor{Gray} Num Layers & Attn. Heads & Hidden size &  FFN Dim \\
 \midrule
  24 & 16 & 1024 & 4096 \\ 
  20 & 20 & 1280 & 3448 \\
  16 & 22 & 1408 & 4096 \\
  12 & 28 & 1792 & 3448 \\
  8 & 36 & 2304 & 3448 \\
  4 & 56 & 3584 & 3448 \\
  \bottomrule
 \end{tabular}
 \label{tab:350m_configs}
% \vspace{-0.5cm}
\end{table}
% \vspace{-10pt}

\section{Simplifying Analytical model} \label{sec:analytical}

\begin{figure}[h]
    \centering
    \includegraphics[width=\linewidth]{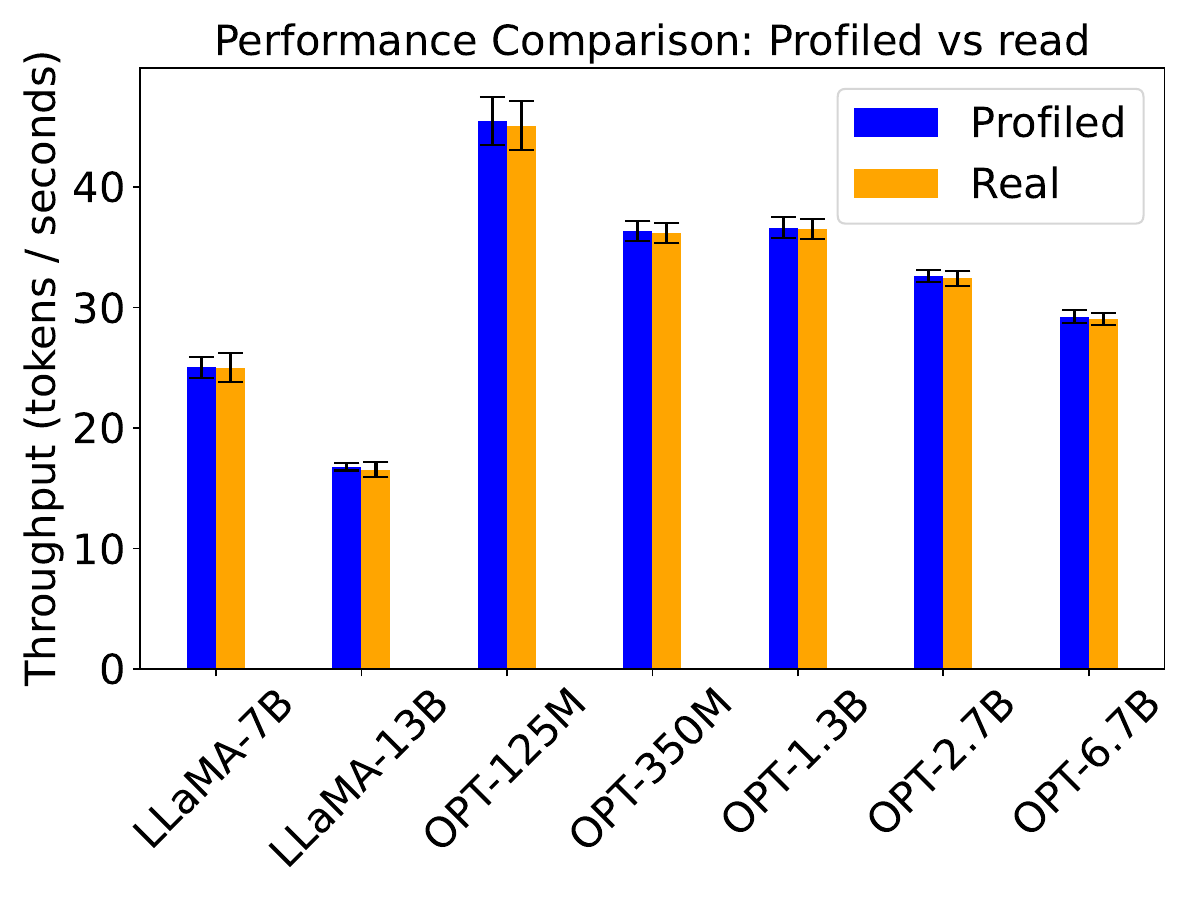}
    \caption{This figure shows that our performance model correctly captures the real performance of speculative decoding. We use \llama-65B and OPT-66B as the target model for each model family, respectively.}
    \label{fig:perf_verification}
\end{figure}

% \minghao{Move to after eval}

\label{sec:analytical_model} 

The original speculative decoding paper~\cite{leviathan2023fast} proposed an analytical model $\frac{1-\alpha^{\gamma + 1}}{(1-\alpha)(\gamma c + 1)}$ to describe the speedup achieved by speculative decoding, where $\alpha$ denotes the expected token acceptance rate (in percentage) and $\gamma$ denotes the look ahead length. However, this model is inaccurate since it assumes that the tokens generated in a sentence are mutually independent. We simplify this cost model and use our updated analytical model in our experiments. 

Assuming a setup similar to prior work~\cite{leviathan2023fast,chen2023accelerating,specinfer} where 
speculative execution of the draft model and target model verification phases happen sequentially, the performance of speculative decoding can be decomposed into the following factors,

\[
% \normalsize
\text{Throughput} =
\begin{cases}
\displaystyle \frac{TAR}{(t^{d}_{target} + t^{d}_{draft})} & \text{if } TAR > 1,\\
\displaystyle \frac{1}{(t^{d}_{target} + t^{d}_{draft})} & \text{if } TAR \leq 1.
\end{cases}
\]

Considering a case where, in each iteration, $d$ tokens are generated by the draft model,
$t^d_{draft}$ depicts the time draft models take to generate $d$ draft tokens, while $t^d_{target}$ is the time taken by the target model for verifying those $d$ draft tokens. 
TAR is used to denote the average number of tokens that were matched across a query or a dataset.

\textbf{Verifying analytical model:} In Figure~\ref{fig:perf_verification}, we compare the throughput predicted by our model with throughput measured on real hardware for two model families: \llama (7B and 13B) and OPT (125M, 350M, 1.3B, 2.7B, and 6.7B) to serve \llama-65B and OPT-66B on MMLU.

We run these experiments on 4 Nvidia 80GB A100 GPUs for 100 iterations on the real server, and the error bars in Figure~\ref{fig:perf_verification} represent the standard deviation of the measurement. For the performance model, we collect $t^d_{draft}$ and  $t^d_{target}$ on a real cluster with a single iteration. For TAR, we collect the average token acceptance rate from the MMLU dataset. The maximum deviation we observed between our proposed analytical model and the results obtained is $3.5\%$.
The close correspondence between our performance model and real measurements shows that our performance model accurately predicts the throughput of speculative decoding. 

\end{document}